 \documentclass[letterpaper, 10 pt, conference]{ieeeconf}
 \IEEEoverridecommandlockouts
\usepackage[bookmarks=true,hidelinks]{hyperref}
\usepackage[utf8]{inputenc}
\usepackage{color}
\usepackage[dvipsnames]{xcolor}
\usepackage{soul}
\usepackage{float}
\usepackage{graphicx}
\graphicspath{ {figures/} }

\usepackage{cite}
\usepackage{tikz}
\usepackage{amssymb}
\usepackage{amsmath}
\usepackage{breqn}
\usepackage{overpic}
\usepackage{resizegather}
\usepackage{siunitx}
\usepackage{afterpage}
\usepackage{textcomp}
\usepackage[export]{adjustbox} 
\usepackage{afterpage} 

\definecolor{myPink}{RGB}{255,229,230}
\definecolor{myOrange}{RGB}{226,145,49}
\definecolor{ployYOrange}{RGB}{253,128,53}
\definecolor{myBlueVec}{RGB}{34,17,247}
\definecolor{myLInfTraj}{RGB}{168,0,14}
\definecolor{myGreenVG}{RGB}{0,160,36}
\definecolor{myDarkerRed}{RGB}{168,0,14}
\definecolor{myPurpleLine}{RGB}{107,0,109}
\definecolor{vt1Blue}{RGB}{95,130,179}
\definecolor{vt3Orange}{RGB}{223,155,52}
\definecolor{myLavender}{RGB}{232,207,229}
\definecolor{myGray}{RGB}{102,102,102}

\newcommand{\norm}[1]{\left\lVert#1\right\rVert}

\usepackage[normalem]{ulem}               

\newcommand\redout{\bgroup\markoverwith
{\textcolor{blue}{\rule[0.5ex]{2pt}{0.8pt}}}\ULon}

\newcommand\blueout{\bgroup\markoverwith
{\textcolor{blue}{\rule[0.5ex]{2pt}{0.8pt}}}\ULon}

\setul{0.5ex}{0.3ex}

\title{\LARGE\bf Minimum-Time Planar Paths with up to Two Constant  Acceleration Inputs and $L_2$ Velocity and Acceleration Constraints}
\author{Victor M. Baez$^{1}$, Haoran Zhao$^{1}$, 
Nihal Abdurahiman$^{2}$, Nikhil V. Navkar$^{2}$,
Aaron T. Becker$^{1}$
\thanks{This work was supported by National Priority Research Program (NPRP) award (NPRP13S-0116-200084) from the Qatar National Research Fund (a member of The Qatar Foundation),
the Alexander von Humboldt Foundation, and the National Science Foundation under  \href{https://www.nsf.gov/awardsearch/showAward?AWD_ID=1932572}{CNS 1932572}, 
\href{https://nsf.gov/awardsearch/showAward?AWD_ID=1849303}{IIS 1849303}, and
\href{https://nsf.gov/awardsearch/showAward?AWD_ID=2130793}{IIS 2130793}. All opinions, findings, conclusions or recommendations expressed in this work are those of the authors and do not necessarily reflect the views of our sponsors.
}
\thanks{$^{1}$ Department of Electrical Engineering, University of Houston, USA, \newline{\tt{ vmontanobaez@gmail.com, atbecker@uh.edu}}}
\thanks{
$^{2}$ Department of Surgery, Hamad Medical Corporation, Doha, Qatar.}}

\date{\today}
\usepackage[switch]{lineno}  

\begin{document}

\maketitle

\begin{abstract}
Given starting and ending positions and velocities, $L_2$ bounds on the acceleration and velocity, and the restriction to no more than two constant control inputs, this paper provides routines to compute the minimal-time path. Closed form solutions are provided for reaching a position in minimum time with and without a velocity bound, and for stopping at the goal position.
 A numeric solver is used to reach a goal position and velocity with no more than two constant control inputs. If a cruising phase at the terminal velocity is needed, this requires solving a non-linear equation with a single parameter.
Code is provided on GitHub\footnote{ \url{https://github.com/RoboticSwarmControl/MinTimeL2pathsConstraints/}}.
\end{abstract}

\section{Introduction and Related Work}

\begin{figure}[tb] 
    \centering
    \begin{overpic}
         [trim={1.6cm 0 0 0},clip,width=0.9\linewidth]{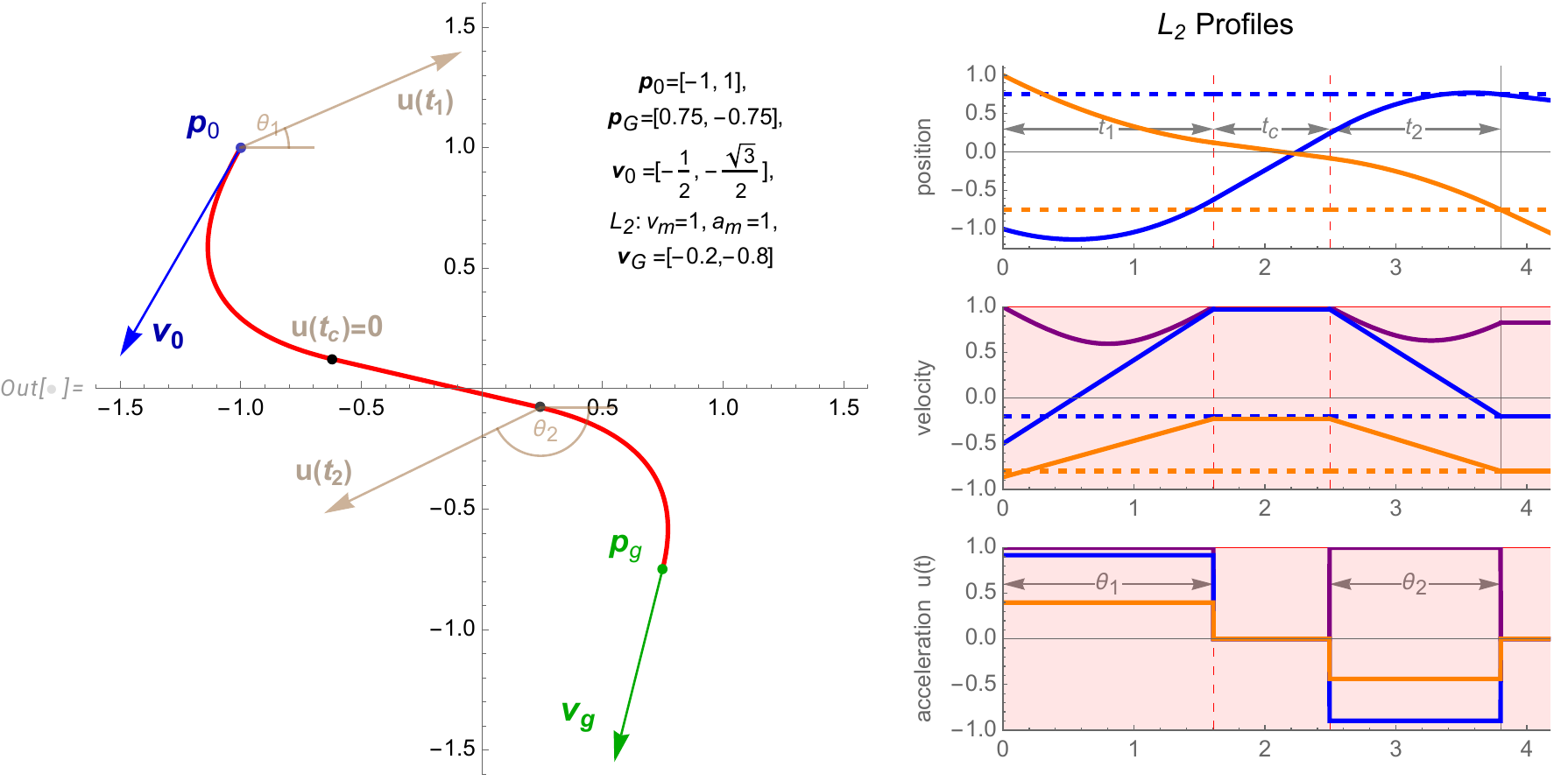}
    \put(77,-1){\tiny time [s]}
    \end{overpic}
    \caption{Left: trajectory of a particle starting from $\textbf{p}_0$ with initial velocity $\textbf{v}_0$ and ending at $\textbf{p}_g$ with ending velocity $\textbf{v}_g$ under two constant acceleration inputs $\textbf{u}$ applied at directions $\theta_1$ and $\theta_2$ for durations $t_1$ and $t_2$; the $\bullet$ shapes show the switching points at $t_1$, $t_c$, and $t_2$ along the path, where $t_c$ is the duration the particle cruises at its maximum velocity $v_m$. Right: $L_2$ position, velocity, and acceleration profiles. \textcolor{myBlueVec}{$x$ in blue}, \textcolor{ployYOrange}{$y$ in orange}, \textcolor{myPurpleLine}{$\sqrt{x^2+y^2}$ in purple}. Bounds on velocity and acceleration are highlighted in pink \textcolor{myPink}{$\blacksquare$}. The control switch times are shown on the position profile.
     See video overview at \href{https://youtu.be/2J-p6CDF4FE}{https://youtu.be/2J-p6CDF4FE}.
        \label{fig:LeadPhoto}
    }
\end{figure}

This paper seeks the minimum-time path for a particle with a restricted set of control inputs: the system can apply no more than two constant thrust inputs, each for a disjoint time.
Moreover, this thrust is limited and there is a constraint on the maximum velocity. 
A representative solution is shown in Fig.~\ref{fig:LeadPhoto}.
We were motivated by a desire for simple optimal control parameterizations of hardware systems with constraints on total maximum acceleration and maximum velocity. 
%
%
%
Many of these problems are currently approximately solved using iterative numeric solvers. 
However, when formulated using $L_2$ bounds, this paper shows there are several problems that provide closed-form solutions, or can be formed as a minimization problem of a single variable. 
The resulting $L_2$ problem is interesting mathematically, and the graphical techniques described in this paper enable an intuitive understanding of the solution. 
This problem could apply to a class of thrusters on a space vehicle such as an astronaut take-me-home system~\cite{duda2018system}, or to other low-friction environments, such as a hovercraft with a single chemical thruster. 

While actuator constraints are often expressed using $L_\infty$ norms, payloads often specify acceleration limits in an $L_2$ sense, such as the 3-G limit on a space shuttle during launch \cite{EvansG-forces}, or an acceleration bound in every direction for translating a cup full of water~\cite{laux2021robot}. 
Similarly, speed limits on highways refer to an $L_2$ speed and not an $L_\infty$ speed.
A given $L_2$ acceleration constraint $a_m$ generates a corresponding $L_\infty$ constraint $a_m/\sqrt{2}$.
This conservative bound can reduce the top acceleration and top speed by almost 30\%.

Restricting the number of control inputs can benefit system performance and longevity, for example by avoiding chatter in systems with high amounts of switching~\cite{de2019mixed,bestehorn2021mixed,ChristianNumSolOptContr2019}. 
From a hardware perspective, limiting the number of control switches can improve lifespans. 
Repeated, alternating stresses are a fundamental concern in mechanical design, particularly when using fatigue-life methods to approximate the lifespan of machine components \cite{machinedesign}. 
Minimizing the number of control switches can facilitate decreasing hardware degradation alongside reducing aggressive actuation by designing
jerk profiles to smoothen trajectories \cite{Lu2017timejerkoptimaltraj},\cite{zhao2015trajectory}.

Optimal paths are common in robotics. 
In 1957, Dubins calculated the shortest planar path for a particle that moved at constant velocity with a constraint on the minimum turning radius~\cite{dubins1957curves}, which has been widely used for planning for mobile robots.
Extending these constraints to acceleration is natural.
Carozza,  Johnson, and Morgan derive the necessary equations for reaching a goal location (and velocity) in minimum time under an acceleration constraint in their paper~\cite{carozza2010baserunner}.
 They show that the fastest $C^{1,1}$ path from one point to another in the plane, given initial velocity, final velocity, and a bound on the magnitude of the acceleration $a_m$, in velocity space is a catenary.
  They applied it to the ``Baserunner's problem'' to determine the sequence of acceleration commands that enables a runner with bounded acceleration to run to all four bases on a baseball field.
  Their numeric process  finds a local minimum.
  They iterate between (a) using prescribed velocities at sequential base positions and optimizing using a  multidimensional Newton’s method with finite difference boundary value methods to determine the path and timing at each baseline, and (b) using a gradient descent method to optimize the velocities at each base.


In general, this is a problem of optimal control with a rich history~\cite{sussmann1997300}.
Numerous approximations have been used. In \cite{TimeOptimalPathsForSoccorBots}, a path between two points with prescribed states is found by generating a maximum allowed velocity profile for a curve described by a spline between the points. 
They find the extremes of their velocity profile using the extremes in the curvature along the spline and hardware limitations on acceleration. 
Afterwards, the spline's control points are optimized for time using parametric programming and a lookup table containing pre-calculated paths.
 Similar work on optimal drone trajectories also used an iterative numeric solver~\cite{hehn2012performance}. 
 Since we directly plan using a small set of acceleration commands, our solution eliminates the need for this iterative procedure.
\subsection{Problem statement}\label{subsec:ProblemStatement}
Given  scalar maximum velocity $v_m$  and acceleration $a_m$, initial position $\mathbf{p}_0 = \mathbf{p}(0)$ and
 velocity $\mathbf{v}_0 = \dot{\mathbf{p}}(0)$ both in $\mathbb{R}^2$, and acceleration equal to the control input
$\ddot{\mathbf{p}}(t) = \mathbf{u}(t)$,
under the constraints that $\Vert \mathbf{u}(t)\Vert_2 \leq a_m$ and $\Vert \dot{\mathbf{p}}(t) \Vert_2 \leq v_m$,
design a $\mathbf{u}(t)$ with a restricted number of changes 
that brings the system to $\mathbf{p}_G = \mathbf{p}(T)$ and
$\mathbf{v}_G = \dot{\mathbf{p}}(T)$, both in $\mathbb{R}^2$, in minimum time $T$. 
For notational convenience, distances are in \si{\meter}, velocities in \si{\meter/\second}, and acceleration in \si{\meter/\second^2}.

%
\subsection[The solution in 1D ]{The solution in 1D  ($L_\infty$ and $L_2$ solutions are equivalent) }\label{subsec:1DSol}

Expressions for time-optimal trajectories for joints of a robot manipulator with velocity and acceleration constraints are provided in \cite{ramos2013time}.
This section uses results from \cite{ramos2013time} to solve the problem in 1D and give context for the remaining sections.

The trajectories vary depending on the initial and final desired states, so velocity profiles are used in \cite{ramos2013time} to classify them as either critical, under-critical, or over-critical, depending on whether the distance $\lvert \textbf{p}_G - \textbf{p}_0 \rvert$ allows for the bound $v_m$ to be reached. The critical profile is defined by a critical displacement, $\Delta p_c$ that results in a linear velocity profile from $\textbf{v}_0$ to $\textbf{v}_G$.
\begin{align}
    \Delta p_c &= s_v \frac{\textbf{v}_G^2 - \textbf{v}_0^2}{2 a_m}, \text{ where $s_v = \textrm{Sign}(\textbf{v}_G - \textbf{v}_0)$}.
\end{align}
%
Without a constraint on velocity, the peak velocity is
\begin{align}
    v_p &=   \sqrt{s_p(p_G - p_0)a_m + \frac{ v_G^2 + v_0^2}{2} }, 
\end{align}
where $s_p = \textrm{Sign}(p_G-p_0 - \Delta p_c)$, $s_p \in \{-1,0,1\}$, accounts for the direction of the initial acceleration.

The acceleration control input for the triangular and trapezoidal velocity profiles are
\begin{align}
     u(t) &=    \left\{
\begin{array}{ll}
\phantom{-}s_p a_m   &  ~~~~~~~~~\,0 \leq t < t_1 \\
\phantom{-}0       &  ~~~~~~~~\,t_1\leq t < t_1+t_c \\
-s_p a_m & ~~~t_1+t_c \leq t < t_1+t_c+t_2 \\
\phantom{-}0 & ~~~~~~~~~\text{otherwise}
\end{array} \right. .
\end{align}


If $v_p \leq v_m$, then the velocity profile will be triangular and 
\begin{align}
    t_1 = \frac{ v_p - s_p v_0}{ a_m}, \quad
    t_c = 0, \quad
    t_2 = \frac{v_p - s_p v_G}{ a_m}.
\end{align}
If $v_p > v_m$, then the velocity profile will be trapezoidal and
\begin{align}
    t_1 &= \frac{ v_m - s_p v_0}{ a_m},\\
    t_c &= \frac{2 s_p a_m  \left(  p_G -p_0 \right) +v_0^2+ v_G^2  - 2  v_m^2 }{2 a_m v_m},\\ 
    t_2 &= \frac{v_m - s_p v_G}{ a_m}.
\end{align}

The total time required is $T = t_1+t_c+t_2$.
This procedure must be modified with more than 1DOF, which
 is more complicated because each DOF must reach the desired position and velocity at the same time.
Kroger and Wahl give a $L_\infty$ solution in \cite{OnlineTrajGen_InstantReactions} that first attempts to make each DOF reach the goal velocity and position at the same time as the slowest DOF.
This is done by solving for a free intermediate cruising speed, which is the root of a sextic polynomial.
  Sometimes a solution in this time is impossible, and a search is conducted to find the next smallest candidate synchronization time.
  
  Similar processes are used for $L_\infty$ controllers with limits on higher derivatives, see~\cite{liu2013time,ata2007optimal,bazaz1999minimum,verscheure2009time,Haoran_IROS2022}.
 We used a related procedure in~\cite{Haoran_IROS2022} and provided open-source code to generate smooth multi-DOF $L_\infty$ trajectories with sinusoidal jerk profiles under jerk, acceleration, and velocity constraints.

\section{Solving for Unconstrained Final Velocity}

What is the fastest way to reach a goal position with one, constant, bounded acceleration input given a starting position and velocity?  
This problem has two variations. Either the goal position is far enough away that the robot reaches maximum velocity and coasts to the goal, or the robot does not reach maximum velocity and accelerates the entire time. 

\begin{figure}[tb]
    \centering
    \begin{overpic}[width=0.48\linewidth]{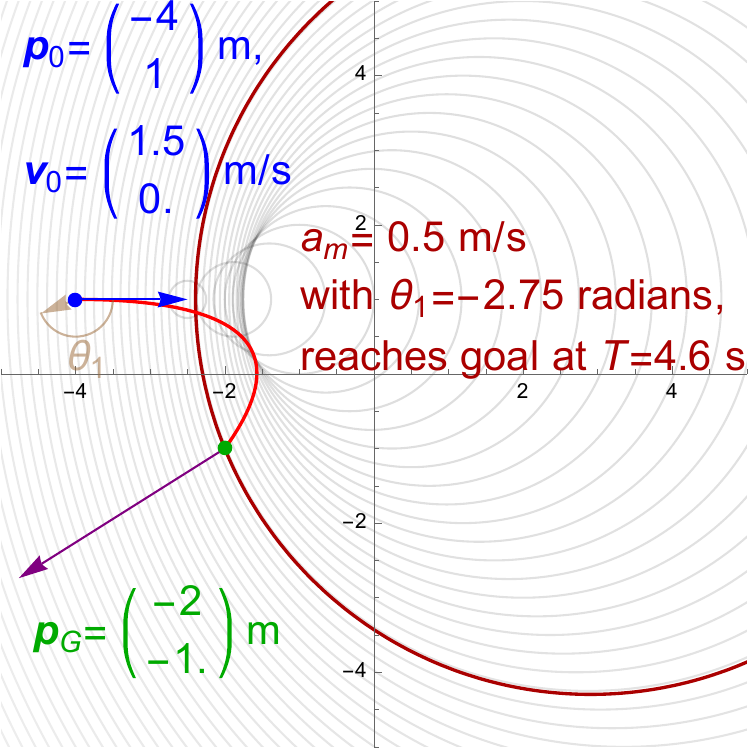}\end{overpic}
    \begin{overpic}[width=0.48\linewidth]{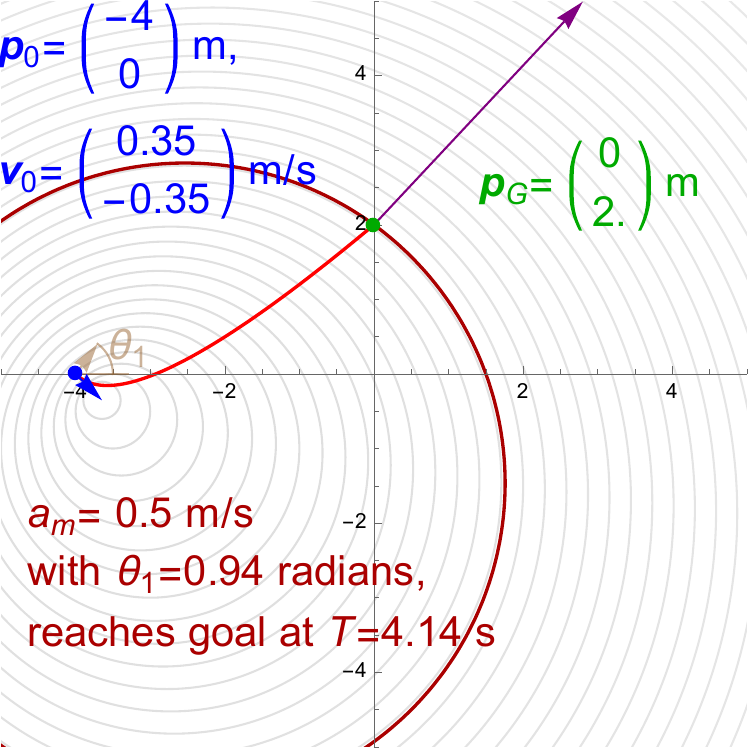}\end{overpic}
    \caption{Two examples of accelerating a particle from a starting position and velocity to a goal position as fast as possible with a bounded input. 
    The locus of reachable positions is a circle whose center moves with $\mathbf{v}_0$. 
     100 isochrones (gray circles \textcolor{gray}{$\bigcirc\!\!\!\!\!\circ$}\,\,) are evenly spaced in squared time:  
$t=  \sqrt{k}$ for $ k \in [0,100]$ to show these loci.  
 }
    \label{fig:ReachPointFastest}
\end{figure}

It is easy to show that all positions are reachable under a constant acceleration input by examining the reachable set.
At time $t$, the locus of positions reachable by the particle is a circle centered at $[p_{0x}+v_{0x}t,p_{0y}+v_{0y}t]$ with radius $\frac{1}{2} a_m t^2$.  The location of the particle on the circle is determined by the angle of acceleration $\theta_1$.
 The gray circles in Fig.~\ref{fig:ReachPointFastest} show these circular loci at different times, and the initial velocity is shown by a blue arrow.

  The particle first achieves the goal position at time $T$. 
  The locus of positions the particle could be  at time $T$ (under all constant accelerations) are drawn in dark red.
  The optimal trajectory is  in red, the optimal constant acceleration input in light brown, and the final velocity on the optimal trajectory is shown with a purple arrow.

The solution has two forms, depending on if the system reaches terminal velocity $v_m$ or not. If it does not, the time $t_1$ can be directly solved and used to solve for $\theta_1$. If the system reaches terminal velocity, finding $t_1$ requires solving for the roots of a sextic equation. The next two sections explain these approaches.
  \subsection[The system does not reach terminal velocity]{The system does not reach terminal velocity $v_m$}
  Without loss of generality, we transform coordinates so that $\mathbf{p}_G$ is the origin.
  If we know the time $t_1$ that the particle reaches $\mathbf{p}_G$, 
we can  solve for the angle of the acceleration:
\begin{align}
   \theta_1(t) = 
 \arctan\left(-\frac{2 (p_{0x} + v_{0x} t_1)}{t_1^2}, \frac{2 (p_{0y}+ v_{0y} t_1)}{t_1^2} \right).
 \end{align}
 Since $t_1 > 0$,  we can simplify this expression to
   $\theta_1(t_1) = \arctan\left(-(p_{0x} + t_1 v_{0x}), p_{0y} + t_1 v_{0y}\right)$.  
The time $t_1$ is when the distance from  $(\textbf{p}_0 + \textbf{v}_0 t_1)$ to $\mathbf{p}_G$ is  $\frac{1}{2} a_m t_1^2$.
Since we translated $\mathbf{p}_G$ to the origin, this results in
\begin{align}
     \left(\frac{1}{2} a_m t_1^2\right)^2 &=  (p_{0x} - v_{0x} t_1)^2 + (p_{0y} - v_{0y} t_1)^2,
\end{align}
  which is quartic in $t_1$.
This is illustrated by the dark red isochrone in Fig.~\ref{fig:ReachPointFastest}.
If we rotate the coordinate frame so $v_{0y} = 0$, and scale velocity 
and positions,
$\tilde{\mathbf{p}}_{0} = \mathbf{p}_{0}/a_m$,
$\tilde{\mathbf{v}}_{0} = \mathbf{v}_{0}/a_m$,
we remove two constants. The smallest non-negative, real $t_1$ is optimal:
\begin{align}
    t_1 &= \{c_4 - c_5, c_4 + c_5, -c_4 + c_5, -c_4 - c_5 \},
\end{align}
where the four roots for $t_1$ are simplified by the coefficients $c_1$ to $c_5$: 
\begin{align}
c_1 &= (9 (\tilde{p}_{0x}^2 - 2 \tilde{p}_{0y}^2) \tilde{v}_{0x}^2 - 2 \tilde{v}_{0x}^6 \nonumber \\
 &+ 
    {\scriptscriptstyle 3 \sqrt{12 (\tilde{p}_{0x}^2 + \tilde{p}_{0y}^2)^3 -  3 (\tilde{p}_{0x}^4 + 20 \tilde{p}_{0x}^2 \tilde{p}_{0y}^2 - 8 \tilde{p}_{0y}^4) \tilde{v}_{0x}^4 + 12 \tilde{p}_{0y}^2 \tilde{v}_{0x}^8}})^{1/3} \nonumber  \\
c_2 &= 2^{4/3} (3 (\tilde{p}_{0x}^2 + \tilde{p}_{0y}^2) - \tilde{v}_{0x}^4)\nonumber \\
c_4 &= \frac{1}{\sqrt{6}}\sqrt{4 \tilde{v}_{0x}^2 + -\frac{c_2}{c_1} + 2^{2/3} c_1}\\
c_3 &= \frac{1}{\sqrt{6}}\left(8 \tilde{v}_{0x}^2 + \frac{c_2}{c_1} - 2^{2/3} c_1 + \frac{12 \sqrt{6} \tilde{p}_{0x}  \tilde{v}_{0x}}{c_4} \right) \nonumber \\
c_5 &= \frac{1}{\sqrt{6}}\left(8 \tilde{v}_{0x}^2 + \frac{c_2}{c_1}  - 2^{2/3} c_1 - \frac{12 \sqrt{6} \tilde{p}_{0x} \tilde{v}_{0x}}{c_4} \right). \nonumber 
\end{align}

The speed of the system at time $t_1$ is $\sqrt{ (v_{0x}+a_m \cos(\theta_1) t_1)^2 + (v_{0y}+a_m \sin(\theta_1) t_1)^2} $.  
If this speed is greater than $v_m$, the system must enter a coasting phase at terminal velocity. 
The solution approach is described in the following section.

\subsection{The system reaches terminal velocity}\label{subsec:reachPointCoasting}

If the ending configuration is sufficiently far from the initial configuration, the goal is reachable in minimum time by a two-phase input which consists of a maximum acceleration input in direction $\theta_1$ for $t_1$ seconds, followed by a coasting phase for $t_c$ seconds.

At time $t_1$ the system reaches velocity $v_m$ under a constant acceleration $a_m [\cos(\theta_1 ),\sin(\theta_1 )]^{\top}$:
\begin{align}
\sqrt{(v_{0x}\! +\! a_m \! \cos(\theta_1 )t_1)^2 \! + \! (v_{0y}\! +\! a_m \! \sin(\theta_1 )t_1)^2} = v_m.
\end{align}
This is a quadratic equation with  two solutions for $t_1$, but only the positive value is relevant since we are planning forward in time.
We express $t_1$ as a function of the angle $\theta_1$:
\begin{align}
   t_1(\theta_1) =&\frac{\sqrt{v_m^2 - v_{0x}^2 - v_{0y}^2 +( v_{0x} \cos(\theta_1 )+ v_{0y} \sin (\theta_1 ))^2 }}{a_m} \nonumber \\
   &- \frac{(v_{0x} \cos (\theta_1)+v_{0y} \sin (\theta_1 ))}{a_m}. \label{eq:t1solutioncoast}
\end{align}

The position of the particle at time $t_1$ is
\begin{align}
p_{x}(t_1) &= p_{0x}+ v_{0x} t_1+\frac{a_m}{2} \cos(\theta_1) t_1^2 \nonumber \\
p_{y}(t_1) &= p_{0y}+ v_{0y} t_1+\frac{a_m}{2} \sin(\theta_1) t_1^2, \label{eq:positionT1}
\end{align}
and the velocity of the particle at time $t_1$ is
\begin{align}
v_{x}(t_1) &= v_{0x} +a_m \cos(\theta_1) t_1 \nonumber \\
v_{y}(t_1) &= v_{0y} +a_m \sin(\theta_1) t_1. \qquad \qquad \label{eq:velocityT1}
\end{align}
\begin{figure}[tb]
    \centering
    \begin{overpic}[width=0.24\linewidth]{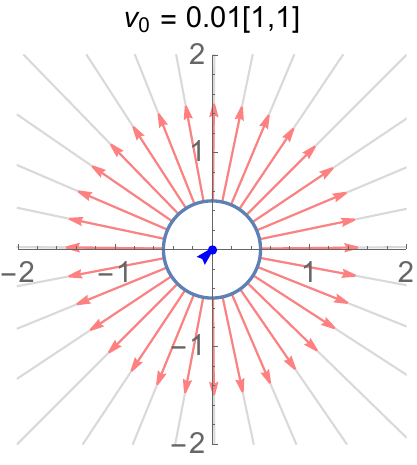}\end{overpic}
    \begin{overpic}[width=0.24\linewidth]{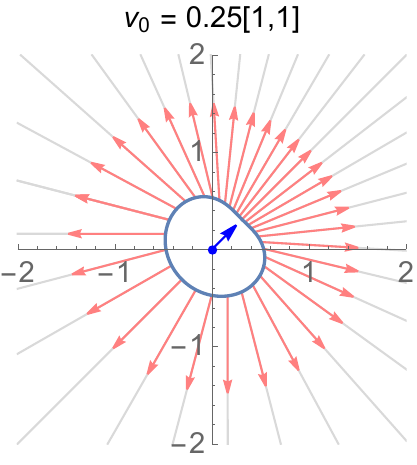}\end{overpic}
    \begin{overpic}[width=0.24\linewidth]{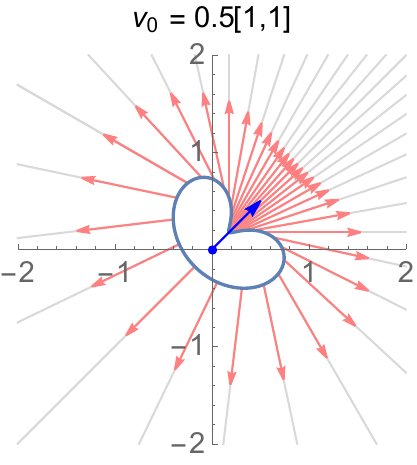}\end{overpic}
    \begin{overpic}[width=0.24\linewidth]{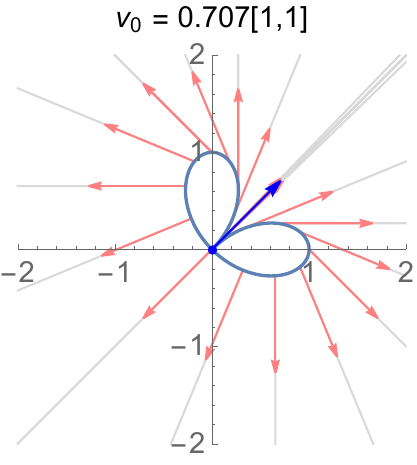}\end{overpic}\\
    \begin{overpic}[width=0.24\linewidth]{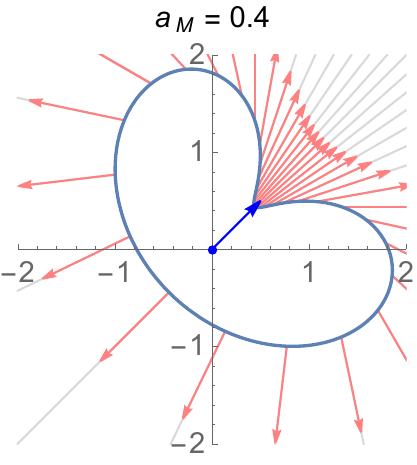}\end{overpic}
    \begin{overpic}[width=0.24\linewidth]{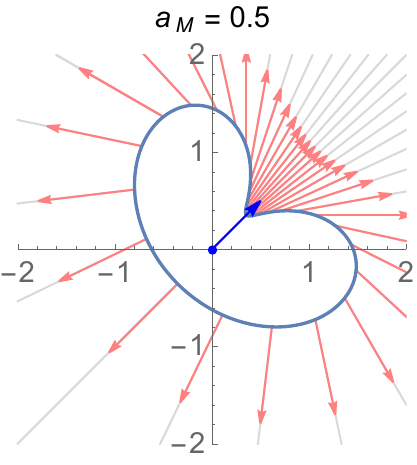}\end{overpic}
    \begin{overpic}[width=0.24\linewidth]{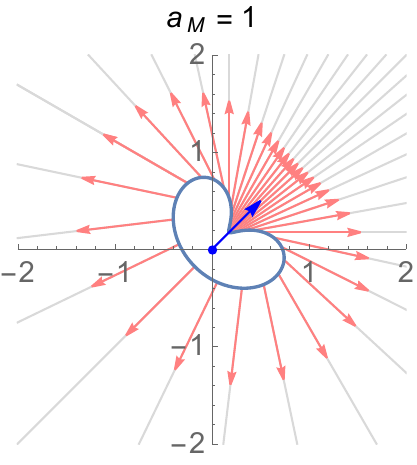}\end{overpic}
    \begin{overpic}[width=0.24\linewidth]{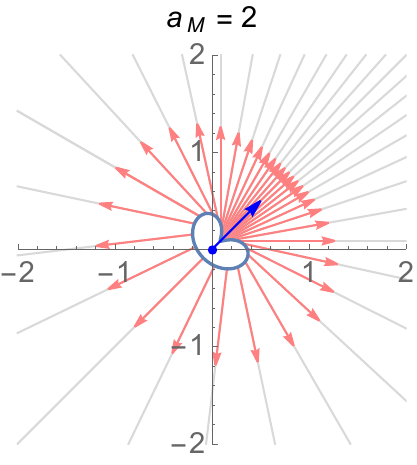}\end{overpic}
    \vspace{-1em}
    \caption{Locus of positions where the particle reaches $v_m = 1$ in light blue \textcolor{vt1Blue}{$\blacksquare$}. Top row shows four different starting velocities (blue).  The velocity of the particle due to   thrust $a_m=1$ in directions $\theta \in k \pi /16$, $k\in [0,31]$ is shown with pink arrows, all of length $v_m$.  These arrows point in every direction.
The bottom row shows $\mathbf{v}_0 = [1/2,1/2]$ for four values of $a_m$. }
    \label{fig:pt1variants}
\end{figure}
Figure~\ref{fig:pt1variants} shows eight variations of the locus of positions at the terminal velocity from \eqref{eq:positionT1} in light blue, along with arrows showing the velocities along this set from \eqref{eq:velocityT1} in pink.
We want solutions for $\theta_1$  that result in the velocity pointing toward to the goal at time $t_1$. 
We could check directly that 
\begin{align}
    \mathrm{arctan}(v_{x}(t_1),v_{y}(t_1)) \equiv \mathrm{arctan}(-p_{x}(t_1),-p_{y}(t_1)), \label{eq:arctanEquiv}
\end{align}
but this involves solving for inverse trigonometric functions.
Instead, we compare the slope of the velocity to the slope of the position error: 
\begin{align}
    \frac{v_{y}(t_1)}{v_{x}(t_1)} \equiv  \frac{-p_{y}(t_1)}{-p_{x}(t_1)}.  \label{eq:slopeForCoast}
\end{align}
This results in two candidate solutions, but we can check both using \eqref{eq:arctanEquiv} and save the correct solution.
We will also have to check for zeros of the equation in the same way.
The resulting equation is
\begin{align}
    v_{x}(t_1)p_{y}(t_1) -v_{y}(t_1) p_{x}(t_1)  \equiv 0.  \label{eq:EqualityCoastToPosition}
\end{align}

Solving for $\sin(\theta_1)$ results in a sextic equation. This equation is long, so it is shared in the Appendix~\cite{baez2023L2eqns}. 
We solve for the six roots of a sextic equation in $\sin(\theta_1) = s$, and discard the complex roots.  
Each remaining root is a solution for $\sin(\theta_1)$ and provides two possible $\theta_1$ solutions since $\theta_1 = \arctan(\pm\sqrt{1-s^2},s)$.
We substitute each possible $\theta_1$ solution into \eqref{eq:t1solutioncoast} to get at most 12 candidate $t_1$ solutions.
The smallest, real, non-negative $t_1$  that satisfies
 \eqref{eq:arctanEquiv}  is used.
 
 Finding the root of a sextic equation can be efficiently computed in many programming languages~\cite{press2007numerical}, for examples see\footnote{\href{https://numpy.org/doc/stable/reference/generated/numpy.roots.html}{https://numpy.org/doc/stable/reference/generated/numpy.roots.html}}$^,$\footnote{\href{https://www.mathworks.com/help/matlab/ref/roots.html}{https://www.mathworks.com/help/matlab/ref/roots.html}}$^,$\footnote{\href{https://www.alglib.net/equations/polynomial.php}{https://www.alglib.net/equations/polynomial.php}}.


\section{Solving for Zero Final Velocity}

\begin{figure}[tb]
    \centering
    \begin{overpic}[width=0.24\linewidth]{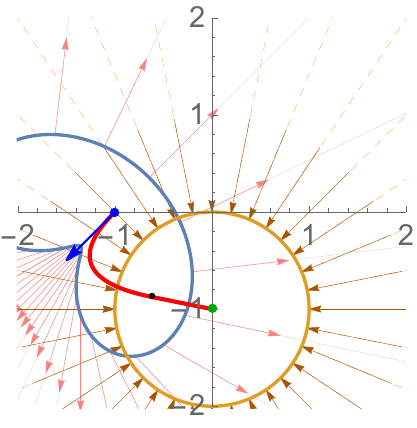}\end{overpic}
    \begin{overpic}[width=0.24\linewidth]{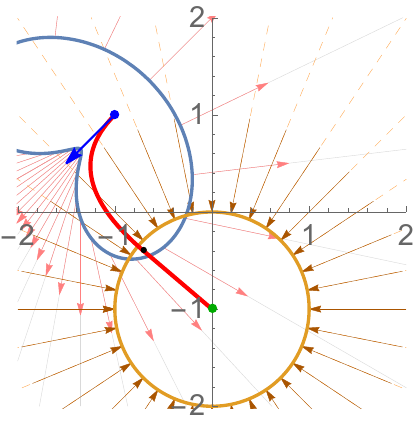}\end{overpic}
    \begin{overpic}[width=0.24\linewidth]{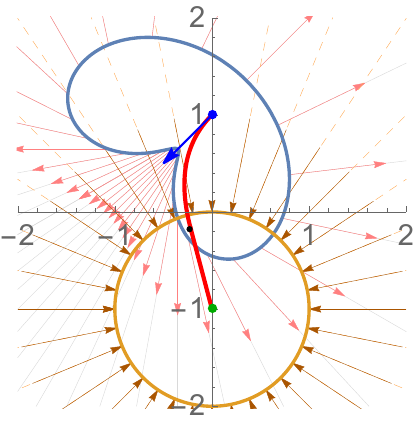}\end{overpic}
    \begin{overpic}[width=0.24\linewidth]{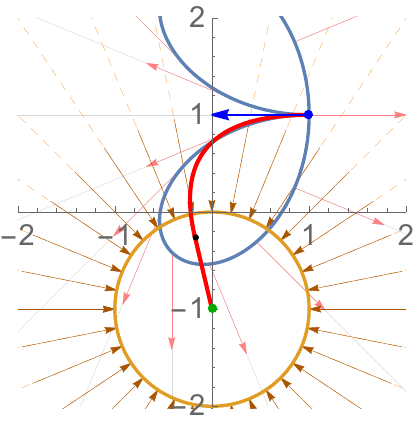}\end{overpic}\\
    \begin{overpic}[width=0.24\linewidth]{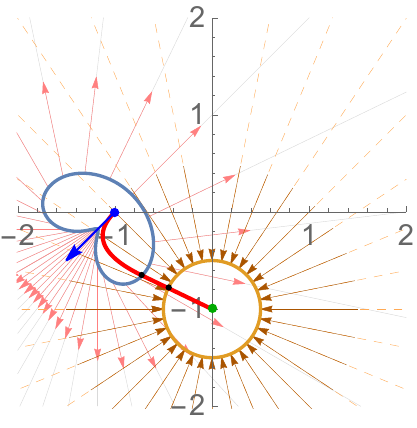}\end{overpic}
    \begin{overpic}[width=0.24\linewidth]{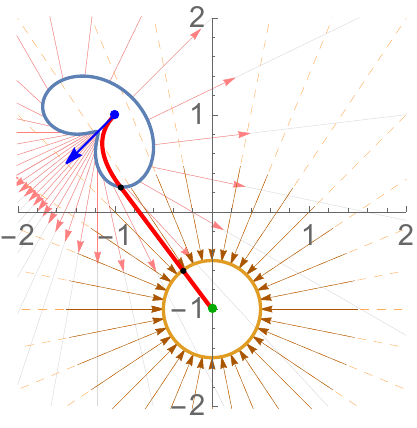}\end{overpic}
    \begin{overpic}[width=0.24\linewidth]{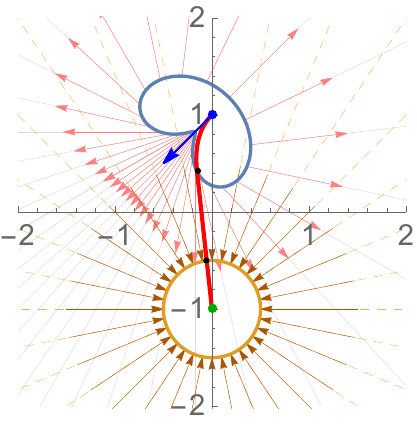}\end{overpic}
    \begin{overpic}[width=0.24\linewidth]{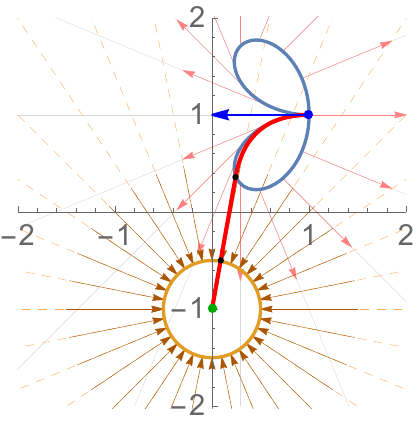}\end{overpic}
    \caption{Solution with nonzero starting and zero ending velocity.  The locus of positions where the particle reaches $v_m = 1$ from the initial position are shown with a light blue set for $\mathbf{p}(t_1)$, \textcolor{vt1Blue}{$\blacksquare$}.  A similar set  constructed starting from the goal position is shown in orange, \textcolor{vt3Orange}{$\blacksquare$}. The solution trajectory is drawn in red.
     The positions where thrust 1 stops and  thrust 2 starts are  indicated by black dots.
 Top row shows four different initial positions with $\mathbf{v}_0 = [-0.5,-0.5]$ or $[-1,0]$ and $a_m = 1/2$. 
 The system never exceeds terminal velocity.
Bottom row shows the same positions and velocities, but with $a_m = 1$, so each requires a coasting phase.\label{fig:zerofinalVelSols}}
\end{figure}

This section provides solutions to problems that stop at the goal position $\mathbf{p}_G$. 
There are two cases depending on if the solution requires coasting at the maximum velocity or not.  Both cases require finding the roots of a sixth order polynomial.
Sample solutions are shown in 
Fig.~\ref{fig:zerofinalVelSols}, which also shows the set $\mathbf{p}(t_1)$ that is calculated with~\eqref{eq:positionT1}, as well as the corresponding set centered at $\mathbf{p}_G$. Because $\mathbf{v}_G = [0,0]$, this set is a circle.

\subsection{Simplifying the problem}
For any goal state with zero velocity, we can transform the coordinates so the goal position is at $[0,0]^{\top}$. We can then rotate the coordinate frame such that $v_{0y} = 0$. 

\subsection{The solution is bang-bang}\label{subsec:ZeroFinalVelNoCoasting}
If the starting and ending position are sufficiently close such that the velocity never exceeds $v_m$, then the goal is reachable in minimum time by a two-phase input which consists of a maximum acceleration input in direction $\theta_1$ for $t_1$ seconds, followed by a maximum acceleration input opposing the current velocity to bring the system to rest in $t_2 = \norm{\mathbf{v}(t_1)}/a_m$ seconds ($t_c = 0$).

 After applying the constant input $a_m[\cos(\theta_1), \sin(\theta_1)]^{\top}$ for $t_1$ seconds, the position and velocity are
 \begin{align}
     \mathbf{p}(t_1) &= \begin{bmatrix}
   & p_{0x} + v_{0x} t_1            + &\frac{a_m}{2} \cos (\theta_1 ) t_1^2\\
   & p_{0y}\phantom{+t_1 v_{0y}} + &\frac{a_m}{2} \sin (\theta_1 ) t_1^2
     \end{bmatrix} \nonumber \\
     \mathbf{v}(t_1) &= \begin{bmatrix}
   & v_{0x}+ & a_m \cos (\theta_1 ) t_1\\
   &         & a_m \sin (\theta_1 ) t_1
     \end{bmatrix}
     \label{eq:PosAndVelTheta1Time1}
 \end{align}
 The deceleration command is in the opposite direction of $\mathbf{v}(t_1)$ so that $\theta_2 = \arctan(-v_{x}(t_1),-v_{y}(t_1))$, and lasts for $t_2 = \norm{\mathbf{v}(t_1)}/a_m $ seconds.
 At time $t_1+t_2$, we want the $x$ and $y$ positions to be zero and the final velocity to be zero.
 The final position is entirely controlled by the initial conditions and the selected $\theta_1$ and $\theta_2$:
 \begin{align}
   t_2 &=\frac{\norm{\mathbf{v}(t_1)}}{a_m} = \sqrt{\left( \frac{v_{0x}}{a_m} + \cos(\theta_1 ) t_1 \right)^2 + (\sin (\theta_1 ) t_1 )^2}\nonumber\\
   0 &= 
  p_{x}(t_1)  + v_{x}(t_1) \frac{t_2}{2}  \nonumber \\
    0 &=
  p_{y}(t_1)  + v_{y}(t_1)  \frac{t_2}{2}. \label{eq:posForZeroFinalVelocity}
 \end{align}
We then scale the starting position and velocity by dividing each by $a_m$ and remove the term $a_m$ from the calculation:  $\tilde{\mathbf{p}}_{0} = \mathbf{p}_{0}/a_m$,
$\tilde{\mathbf{v}}_{0} = \mathbf{v}_{0}/a_m$.
 We apply a change of variables to eliminate the two trigonometric functions: $\cos (\theta_1 )= c$, and $\sin (\theta_1)=\pm \sqrt{1-c^2}$.
 The resulting position constraints simplify to:
 \begin{align}
    0 &= 2 \tilde{p}_{0x} + 2 \tilde{v}_{0x} t_1\! +c t_1^2 + (\tilde{v}_{0x}\! +\! c t_1) \sqrt{\tilde{v}_{0x}^2 \!+\! 2 c  \tilde{v}_{0x} t_1 +\!t_1^2} \nonumber\\
    0 &= 2 \tilde{p}_{0y} + \sqrt{1-c^2} t_1 \!\left(\!\sqrt{\tilde{v}_{0x}^2 + 2 c  \tilde{v}_{0x} t_1 +\! t_1^2}+\!t_1\!\right).\! \label{eq:positionUsingThrustToStopAtGoalNoTermalVelocity}
 \end{align}
 This set of equations can be solved for $c$ as a function of $t_1$.
The calculations are long, and are shared in Appendix~\cite{baez2023L2eqns} (see also the code implementation).


We solve for the roots of this sextic equation in $t_1$ to get six candidate $t_1$ values. 
We substitute each non-negative, real value into a closed-form equation to compute candidate $c= \cos(\theta_1)$ from $t_1$.  Since $\theta_1 = \arctan(c,\pm\sqrt{1-c^2})$, this provides at most 12 candidate $\theta_1$ values. 
We test each $\theta_1$ value in \eqref{eq:posForZeroFinalVelocity} and select the solution with zero position error that minimizes the total time.

Figure~\ref{fig:zerofinalVelSols}, top row shows four solutions.  In each, the solution trajectory is entirely contained within the union of the $\mathbf{p}(t_1)$ locus and a maximum braking radius circle centered at $\mathbf{p}_G$. The switching position at time $t_1$ is shown with a black point.

\subsection{The solution requires a cruising phase}

If the solution from Section \ref{subsec:ZeroFinalVelNoCoasting} requires a velocity larger than $v_m$, we must have a cruising phase at the maximum velocity.
 The goal is reachable by a three-phase input.
 This control consists of a maximum acceleration input in direction $\theta_1$ for $t_1$ seconds, followed by a cruising phase for $t_c$ seconds, followed by a maximum acceleration input opposing the current velocity to bring the system to rest in $t_2$ seconds.
 
 We first find the two-phase solution from Sec.~\ref{subsec:reachPointCoasting} to reach the goal position $\mathbf{p}_G$ by accelerating in direction $\theta_1$ for $t_1$ seconds.  
 Rather than cruising from time $t_1$ to the goal, we need to start braking when we are $t_2$ seconds away from the goal.
 \begin{subequations}
 \begin{align}
     r &= \frac{v_m^2}{2 a_m} \text{\quad braking radius.} \label{eq:brakingRadius}
     \\ 
     &\text{ distance from $\mathbf{p}(t_1)$ to goal:} \nonumber\\
d &= \norm{  \mathbf{p}_G - \left(\mathbf{p}_0 +  \mathbf{v}_0 t_1 + \frac{a_m}{2}  \begin{bmatrix} \cos(\theta_1)\\ \sin(\theta_1)\end{bmatrix} t_1 ^2\right)
   }\\
t_c &= \frac{d - r}{v_m} \text{ cruising time,  } 
t_2 = \frac{v_m}{a_m} \text{\quad braking time.} \\ 
&\text{ braking direction:} \\
\theta_2 &= -\arctan\left(v_{0x}+a_m \cos(\theta_1) t_1 , v_{0y}+a_m \sin(\theta_1) t_1 \right). \nonumber
 \end{align} \label{eq:stoppingCoastingSolution}
  \end{subequations}
The thrust time $t_1$ is a function of $\theta_1$ as given in \eqref{eq:t1solutioncoast}, and so
 \eqref{eq:stoppingCoastingSolution} are all functions of only $\theta_1$.  

The bottom row of Fig.~\ref{fig:zerofinalVelSols} shows four representative solutions.  The $\mathbf{p}(t_1)$ loci are shown in light blue, and the corresponding set backwards from $\mathbf{p}_G$ is a golden-colored circle with radius~\eqref{eq:brakingRadius}. Two black points show where the first and the second thrust commands are applied along the solution trajectory.

\section{Solving for Non-Zero Final Velocity}

Solving for a non-zero final velocity $\mathbf{v}_G$ is harder, but is necessary for smoothly traversing through waypoints~\cite{lin2018efficient}.  A final velocity adds two parameters to the equations. 
Currently, we use a numeric solver to find a bang-bang solution to get to the goal with no velocity limit.    If this solution results in a maximum velocity greater than $v_m$ at the switching point of the bang-bang controller, we call a second function to solve for a cruising phase.

\begin{figure}[tb]
    \centering
        \begin{overpic}[width=0.24\linewidth]{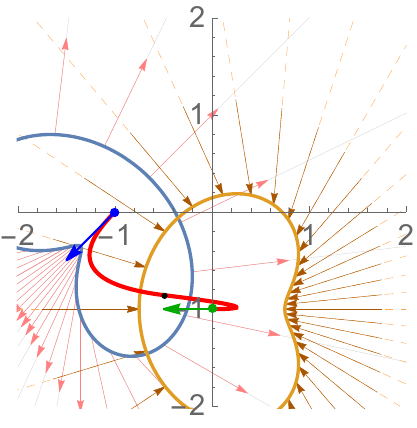}\end{overpic}
    \begin{overpic}[width=0.24\linewidth]{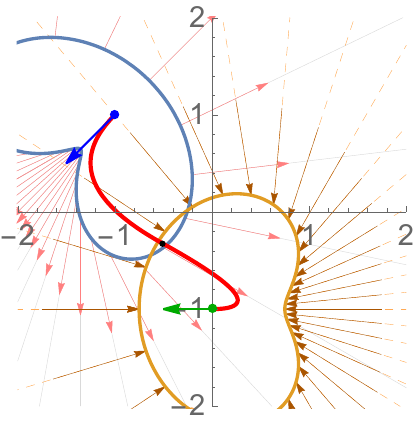}\end{overpic}
    \begin{overpic}[width=0.24\linewidth]{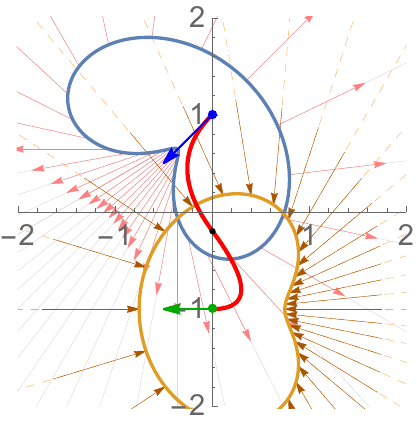}\end{overpic}
    \begin{overpic}[width=0.24\linewidth]{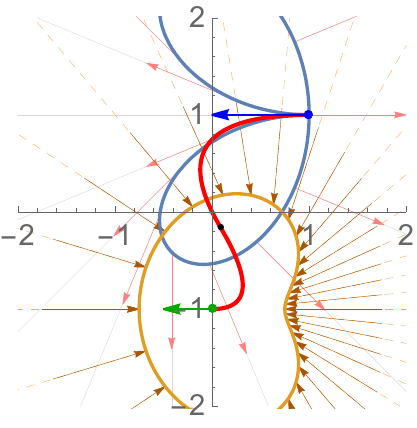}\end{overpic}\\
    \begin{overpic}[width=0.24\linewidth]{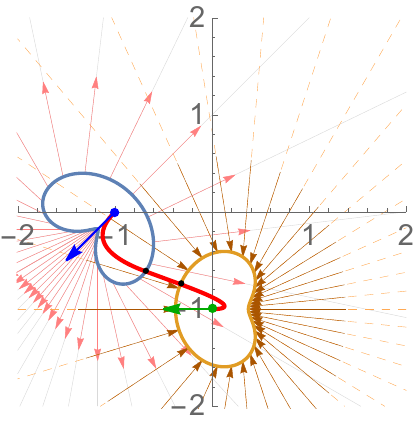}\end{overpic}
    \begin{overpic}[width=0.24\linewidth]{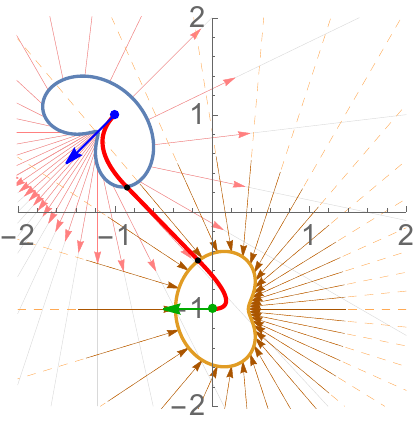}\end{overpic}
    \begin{overpic}[width=0.24\linewidth]{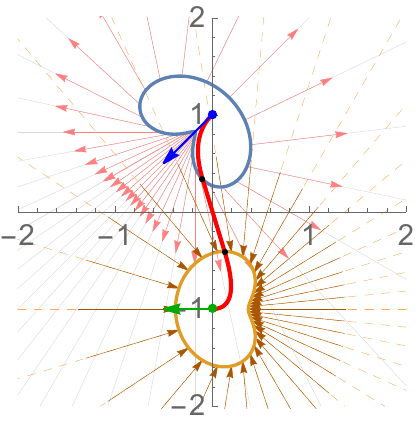}\end{overpic}
    \begin{overpic}[width=0.24\linewidth]{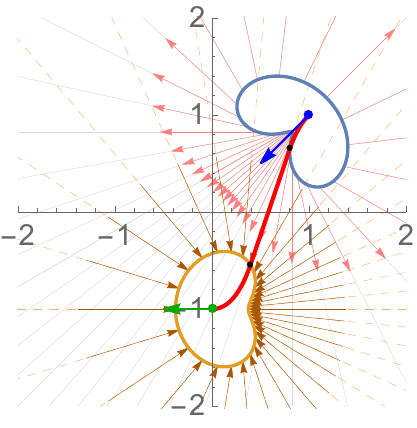}\end{overpic}
    \vspace{-1em}
    \caption{Solution with nonzero starting and ending velocity.  The locus of positions where the particle reaches $v_m = 1$ from the initial position are shown with a light blue set for $\mathbf{p}(t_1)$.  A similar set  constructed starting from the goal position using $-\mathbf{v}_G$ is shown in orange. The solution trajectory is drawn in red.
     The positions where thrust 1 stops and  thrust 2 starts are  indicated by black dots.
 Top row shows four different initial positions with $\mathbf{v}_0 = [-0.5,-0.5]$, $\mathbf{v}_G = [-0.5,0]$ and $a_m = 1/2$ (the right differs to  ensure $v_m$ is not reached).  The system never reaches terminal velocity. 
Bottom row shows the same positions and velocities, but with $a_m = 1$. The system reaches terminal velocity $v_m$ in each case.
\label{fig:finalVelSols}
}
\end{figure}

Sample solutions are shown in 
Fig.~\ref{fig:finalVelSols}, which also shows the set $\mathbf{p}(t_1)$ that is calculated with~\eqref{eq:positionT1}.  The corresponding set  centered at $\mathbf{p}_G$ is generated using the same process, but with $-\mathbf{v}_G$ as input.

\subsection{No cruising phase, non-zero final velocity}\label{subsection:NoCoastNonzeroFinal}

The bang-bang controller is described by the following inputs, which apply maximum thrust in direction $\theta_1$ for $t_1$ seconds, and then maximum thrust in direction $\theta_2$ for $t_2$ seconds. There is no cruising phase, so $t_c = 0$. 
Here, $\mathbf{p}(t_1)$ and $\mathbf{v}(t_1)$ are the same as in (\ref{eq:positionT1}) and (\ref{eq:velocityT1}). 
Define the position and velocity generated by starting at $\mathbf{p}_G$ with velocity $\mathbf{v}_G$ and applying acceleration in direction $\theta_2$  \emph{backwards in time} for $t_2$ seconds as $[\mathbf{p}_{\text{-}t_2}, \mathbf{v}_{\text{-}t_2}]$. 
\begin{subequations}
\begin{align}
%
p_{x}(t_1)& \equiv
p_{Gx} - v_{Gx}  t_2  +\frac{a_m}{2}      \cos(\theta_2) t_2^2 & =p_{\text{-}t_2,x} \\
p_{y}(t_1)& \equiv
p_{Gy} - v_{Gy}  t_2 +\frac{a_m}{2}      \sin(\theta_2) t_2^2 & =p_{\text{-}t_2,y}\\
%
%
v_{x}(t_1) & \equiv
v_{Gx} - a_m  \cos(\theta_2) t_2 & =v_{\text{-}t_2,x}\\ 
v_{y}(t_1) & \equiv
v_{Gy} - a_m  \sin(\theta_2) t_2 & =v_{\text{-}t_2,y}
\end{align}
\label{eq:posAndVelaftert1ANDbeforet3}
\end{subequations}
Solving this nonlinear set of constraints requires a good starting estimate.
We run the solver multiple times, using a set of candidate guesses for the unknowns $[\theta_1, t_1,\theta_2, t_2].$  
To  generate good candidate guesses, we first wrap the procedure in Section~\ref{subsec:ZeroFinalVelNoCoasting} into the function 
\begin{align}
    (\theta_1,t_1) = \mathrm{stopAtGoalNoCoast}[ \mathbf{p}_0,\mathbf{v}_0,\mathbf{p}_G,a_m,v_m ],
\end{align} that returns the necessary thrust direction and time for a given initial position and velocity,  a desired stopping position $\mathbf{p}_G$, and acceleration and velocity constraints.
Next, we generate a set of positions $pts$:
\begin{align}
pts &= \{\mathbf{p}_0,~ \mathbf{p}_1 , ~ \frac{1}{2}\left(\mathbf{p}_1 +\mathbf{p}_4 \right), ~ \mathbf{p}_4
   ,~ \mathbf{p}_G \}, \text{ where}\\
\mathbf{p}_1 &= \mathbf{p}_0 + \frac{\norm{\mathbf{v}_0}}{2 a_m} \mathbf{v}_0, \quad
\mathbf{p}_4 = \mathbf{p}_G -  \frac{\norm{\mathbf{v}_G}}{2 a_m}. 
\end{align}
Here  $\mathbf{p}_1$ is reached by maximum braking starting from $\mathbf{p}_0$ with velocity $\mathbf{v}_0$, and $\mathbf{p}_4$ is reached by maximum braking from  $\mathbf{p}_G$ and initial velocity $-\mathbf{v}_G$. 
We use candidate starting values 
\begin{align}
 \mathrm{stopAtGoalNoCoast}[ \mathbf{p}_0,\mathbf{v}_0,pts[k],a_m,v_m ], ~ k \in \{3,4,5\}\nonumber \\
 \mathrm{stopAtGoalNoCoast}[ \mathbf{p}_G,-\mathbf{v}_G,pts[j],a_m,v_m ], ~ j \in \{1,2,3\}\nonumber  
\end{align}
Of all solutions that reach the goal, we select the solution that minimizes the total time.
The fastest velocity occurs at the switching time.  If $\norm{\mathbf{v}_{t_1}} > v_m$ a cruising phase is required.

\subsection{Cruising phase, non-zero final velocity}
If the solution from Sec.~\ref{subsection:NoCoastNonzeroFinal} requires a velocity greater than $v_m$, we must have a cruising phase at the maximum velocity.
Solving with a cruising phase is conceptually easier.
We know that the solution trajectory reaches terminal velocity, and thus $t_1$ is determined by $\theta_1$ and $t_2$ by $\theta_2$, both by using~\eqref{eq:t1solutioncoast}.
We merely need to find a $\theta_1$ and a $\theta_2$ that solve the problem.
The velocities while cruising and the scaled difference in position are all equal, and along some unknown vector $\phi$:
\begin{align}
    \mathbf{v}_{t_1} \equiv  \mathbf{v}_{\text{-}t_2} \equiv v_m \frac{\mathbf{p}_{\text{-}t_2} - \mathbf{p}_{t_1}}{\norm{\mathbf{p}_{\text{-}t_2} - \mathbf{p}_{t_1}} } \equiv  v_m \begin{bmatrix} \cos{(\phi)} \\ \sin{(\phi)}
    \end{bmatrix}.
\end{align}
Given a $\phi$, we can solve for $\theta_1$ and  $\theta_2$:
\begin{align}
   \theta_1 &= \arctan(v_m \cos{(\phi)} - v_{0x}, v_m \sin{(\phi)} - v_{0y}  ) \\
   \theta_2 &= \arctan(v_{Gx} - v_m \cos{(\phi)}, v_{Gy} - v_m \sin{(\phi)}  )
\end{align}
Then we must solve for the parameter $\phi$:
\begin{align}
    \phi \equiv & \arctan( \mathbf{p}_{\text{-}t_2x} - \mathbf{p}_{t_1x}, 
     \mathbf{p}_{\text{-}t_2y} - \mathbf{p}_{t_1y}), \text{ let}\nonumber \\
     r_0 =& \sqrt{(v_{0x}-v_m \cos (\phi ))^2+(v_{0y}-v_m \sin (\phi ))^2}, \nonumber \\
     r_G =& \sqrt{(v_{Gx}-v_m \cos (\phi ))^2+(v_{Gy}-v_m \sin (\phi ))^2}, \text{ then} \nonumber \\
    \phi \equiv & \scriptstyle \arctan(2 a_m (p_{0x}+p_{Gx})-v_m (r_0+r_G) \cos (\phi )-r_0 v_{0x}-r_G v_{Gx},\nonumber \\
    & \scriptstyle-2 a_m (p_{0y}-p_{Gy})-v_m (r_0+r_G) \sin (\phi )-r_0 v_{0y}-r_G v_{Gy}).
\end{align}
This equation is nonlinear, but $\phi$ is the only variable. 
We use a Van der Corput sequence\footnote{\url{https://rosettacode.org/wiki/Van_der_Corput_sequence}} with base 2 in the range $[-\pi, \pi]$ to sample $\phi$ evenly with increasing refinement. We then perform a root finding operation on $\phi$, initializing our guess with an element from the Van der Corput sequence, and iterating until the $\norm{\mathbf{p}_G - \mathbf{p}(T)} + \norm{\mathbf{v}_G - \mathbf{v}(T)}$ is below a bound $e_{\min}$.

To test this algorithm, we randomly generated initial and final positions $(\mathbf{p}_{0},\mathbf{p}_{G})$ within a circle of radius $r=2$, and velocities $(\mathbf{v}_0,\mathbf{v}_G)$ in a radius $r=1$ until we had 10,000 initial conditions that required a cruising phase for $a_m = 1, v_m = 1$. We used $e_{\min} = 10^{-12}$.
Of these, 9562 converged within $e_{\min}$ using the first sequence value, 9712 in the first two values, and 9911 in the first 5 values. The average number of values needed was 1.17, and the largest was 87.
Solving 10,000 queries required 26.5 seconds on 3.3 GHz i7 laptop.

\section{Trajectory Examples and Analysis}

This section showcases several examples and contrasts our minimum-time trajectory solutions modeled using $L_2$ bounds to those modeled with $L_\infty$ bounds. 
The time-optimal trajectories modeled with $L_\infty$ bounds were calculated using methods in \cite{ramos2013time}, and the two DOF’s were synchronized using Kroger and Wahl’s search technique from \cite{OnlineTrajGen_InstantReactions}, as described in Sec.~\ref{subsec:1DSol}. 

\begin{figure}[ht] 
    \centering
    \includegraphics[width=1\linewidth]{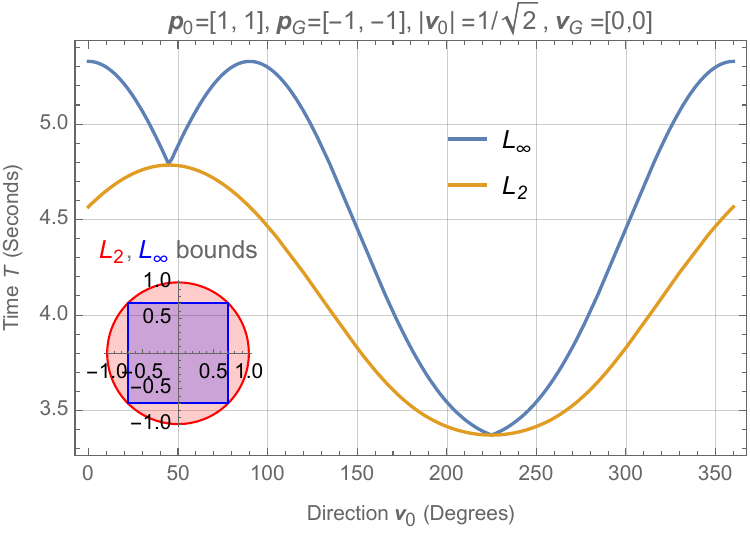}
    \caption{Comparing the finishing times for $L_2$ and $L_\infty$ solutions with bounds $L_2: a_m = 1, v_m = 1$ and $L_{\infty}: a_m = 1/\sqrt{2}, v_m = 1/\sqrt{2}$, which are shown in the inset graphic in the lower left. 
    Because of these bounds the $L_{\infty}$ solution is in general longer than the $L_{2}$ solution.
    \label{fig:timeForLinfAndL2comparison} }
     \vspace{-1em}
\end{figure}

\begin{figure}[ht] 
    \centering
    \includegraphics[width=1\linewidth]{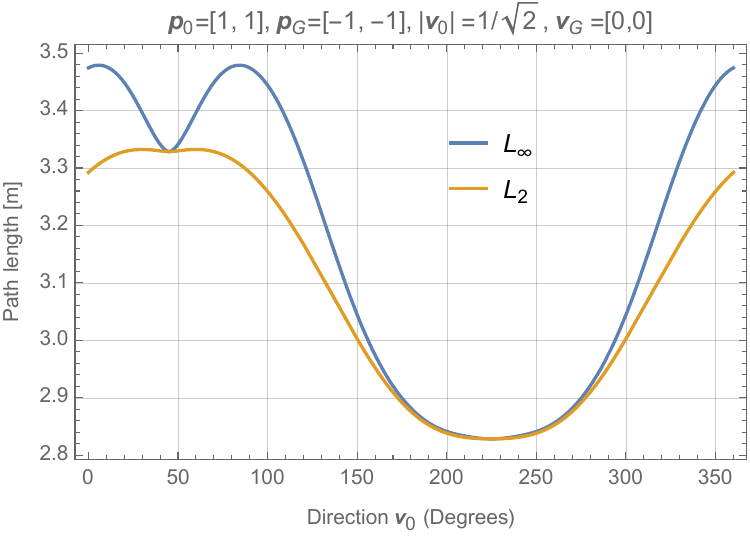}
    \caption{Comparing the total path length for $L_2$ and $L_\infty$ solutions with bounds $L_2: a_m = 1, v_m = 1$ and $L_{\infty}: a_m = 1/\sqrt{2}, v_m = 1/\sqrt{2}$. 
    \label{fig:pathLengths}}
     \vspace{-1em}
\end{figure}

Using a controller designed using $L_2$ bounds results in a controller that is never slower than a control that uses the $L_{\infty}$ bounds that obey the $L_2$ bounds. 
The times are only the same if the solution trajectory lies entirely along a slope of $\pm 1$. 
This is illustrated in Fig.~\ref{fig:timeForLinfAndL2comparison} for $\textbf{p}_0=[1,1]$, $\textbf{p}_G=[-1,-1]$;  finishing times for $L_2$ and $L_\infty$ are only equal for $\textbf{v}_0$ angular directions of \SI{45}{\degree} and \SI{225}{\degree}. At these $\textbf{v}_0$ directions, the problem is effectively 1D and the $x$ and $y$ velocities are identical. Fig. \ref{fig:pathLengths} shows the path lengths for the same conditions.

\begin{figure*}[ht]
    \centering    
    \begin{overpic}[trim={1.6cm 0 0 0},clip,width=0.9\linewidth]{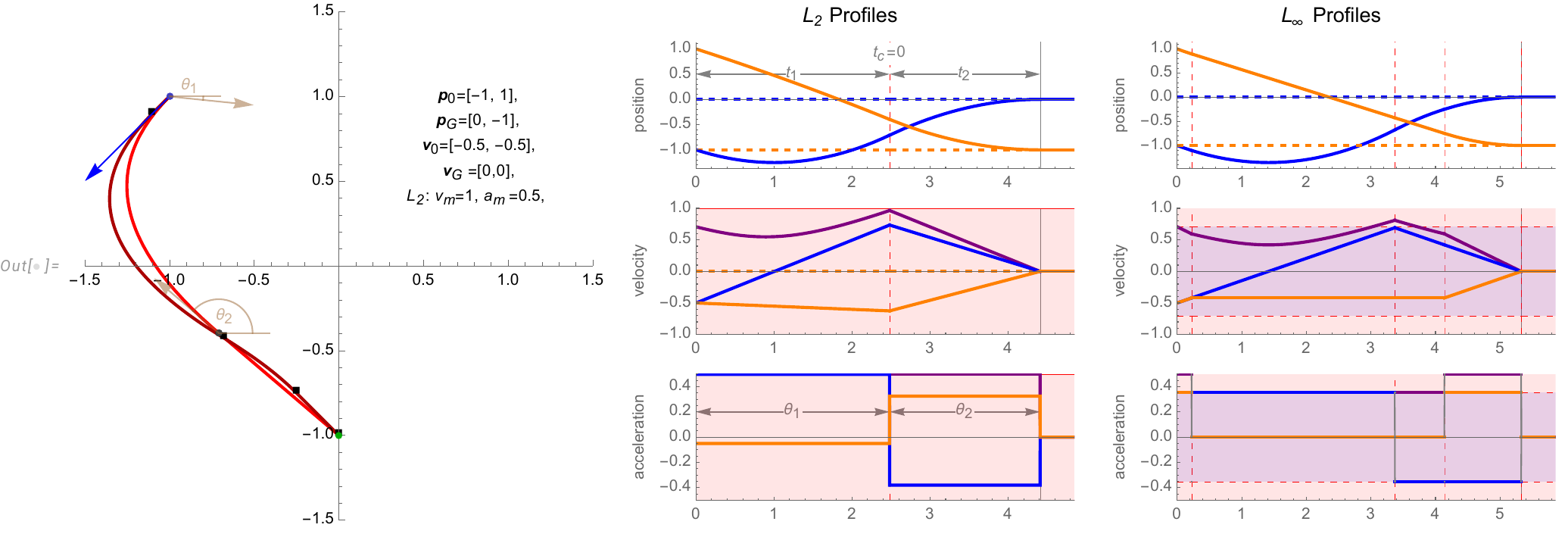}
        \put(0,32){\textcolor{black}{\textbf{(a)}}}
        \put(-5,5){\small\rotatebox{90}{\parbox{4cm}{Stopping at goal,\\no cruising phase in $L_2$. }}}
        \end{overpic}
    \begin{overpic}[trim={1.6cm 0 0 0},clip,width=0.9\linewidth]{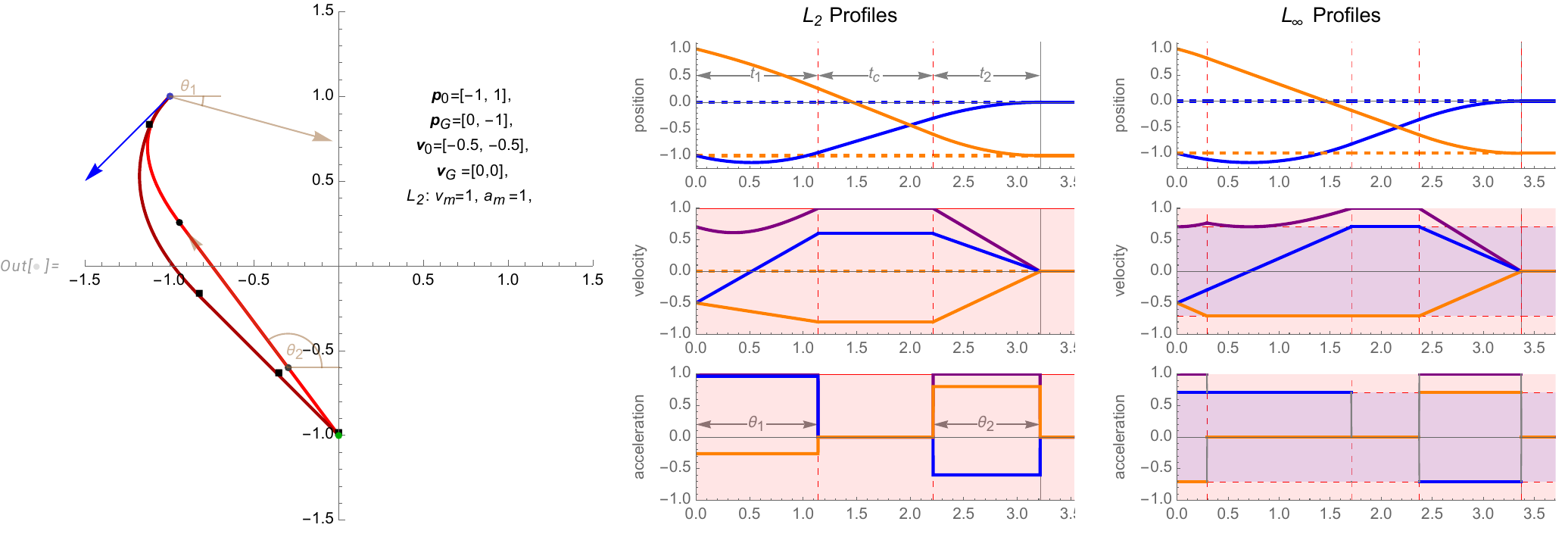}
        \put(0,32){\textcolor{black}{\textbf{(b)}}}
        \put(-5,5){\small\rotatebox{90}{\parbox{4cm}{Stopping at goal,\\with cruising phase in $L_2$. }}}
        \end{overpic}
    \begin{overpic}[trim={1.6cm 0 0 0},clip,width=0.9\linewidth]{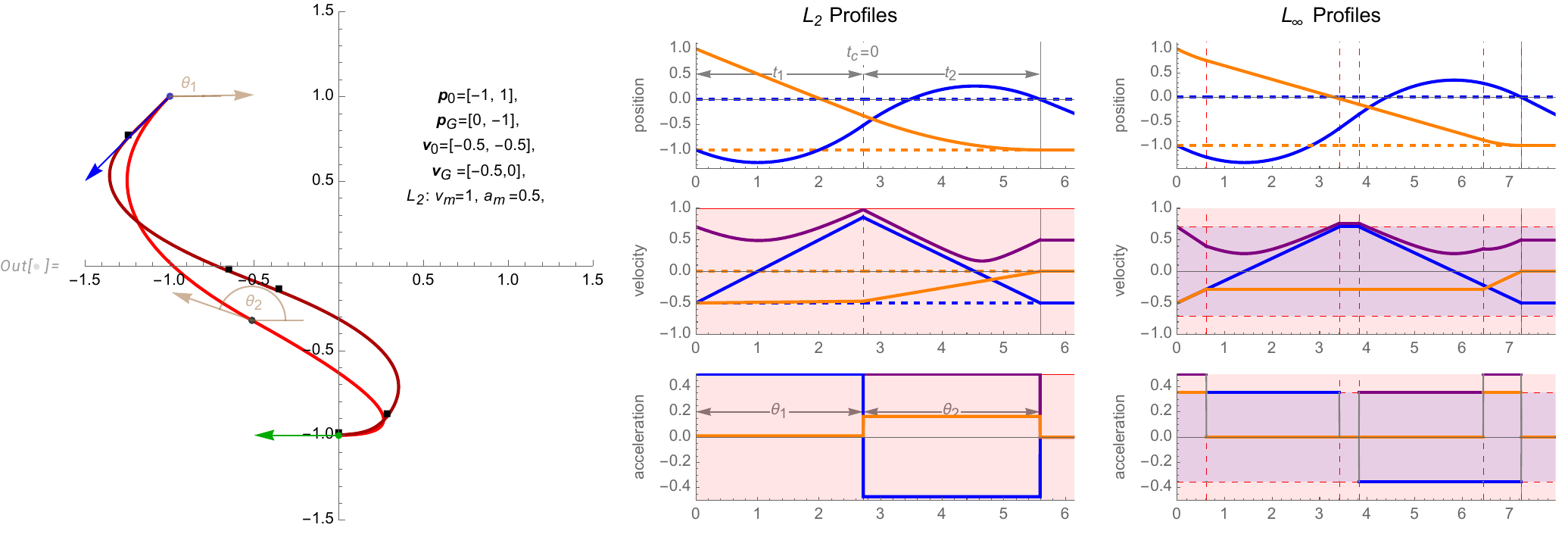}
        \put(0,32){\textcolor{black}{\textbf{(c)}}}
        \put(-5,5){\small\rotatebox{90}{\parbox{4cm}{Non-zero goal velocity,\\no cruising in $L_2$. }}}
        \end{overpic}
   \begin{overpic}[trim={1.6cm 0 0 0},clip,width=0.9\linewidth]{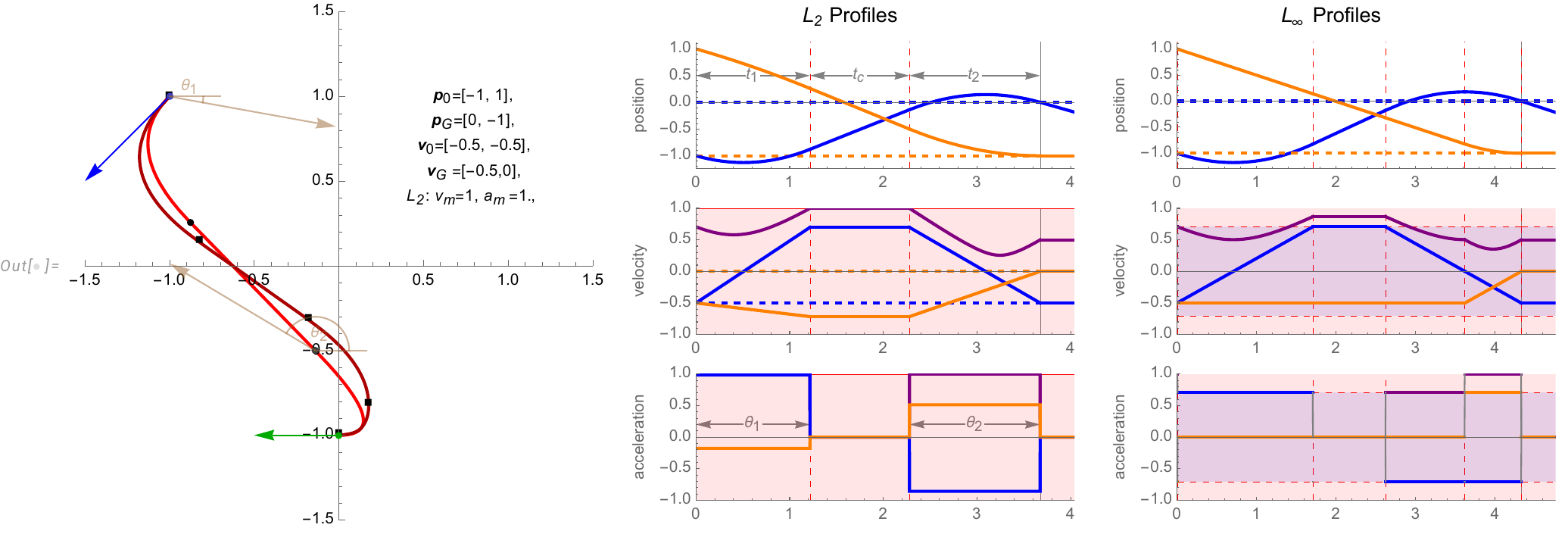}
        \put(0,32){\textcolor{black}{\textbf{(d)}}}
        \put(-5,5){\small\rotatebox{90}{\parbox{4cm}{Non-zero goal velocity,\\with cruising phase in $L_2$. }}}
        \put(53.5,-1){\scriptsize \textcolor{myGray}{\tiny time [s]}}
        \put(85,-1){\scriptsize \textcolor{myGray}{\tiny time [s]}}
        \end{overpic}
    \caption{   
        Left column: trajectory plots for \textcolor{red}{$L_2$ in red} and \textcolor{myDarkerRed}{$L_{\infty}$ in dark red}. 
        Control switches are shown by $\bullet$ shapes along the $L_2$ path and {\tiny$\blacksquare$} shapes along the $L_\infty$ path.
        Right column: $L_2$ and $L_{\infty}$ position, velocity, and acceleration profiles.
    \textcolor{myBlueVec}{$x$ in blue}, \textcolor{ployYOrange}{$y$ in orange}, and \textcolor{myPurpleLine}{$\sqrt{x^2+y^2}$ in purple}.
    Bounds for $L_2$ are in pink \textcolor{myPink}{$\blacksquare$} and for $L_{\infty}$ in lavender \textcolor{myLavender}{$\blacksquare$}.
    \label{fig:trajsAndProfiles}}
    \vspace{-1em}
    \end{figure*}

The plots on the left of Fig.~\ref{fig:trajsAndProfiles} (a) through (d)  show our $L_2$  paths in red and $L_\infty$  bounded paths in dark red for several cases. 
Each of these figures also include three plots on the right that show the position, velocity, and acceleration profiles for each case. The highlighted pink areas in the velocity and acceleration profiles show the $L_2$ bounds. The corresponding $L_\infty$ bounds are highlighted in lavender.

Figures~\ref{fig:trajsAndProfiles}(a) and ~\ref{fig:trajsAndProfiles}(b) show solutions for cases that stop at the goal position given a starting velocity, $\mathbf{v}_0$.
Fig.~\ref{fig:trajsAndProfiles}(a) shows solutions for when no cruising phase is required, while Fig.~\ref{fig:trajsAndProfiles}(b) shows the solutions for a case when a cruising phase is required. 
Solutions with a specified goal velocity for are shown in Fig.~\ref{fig:trajsAndProfiles}(c) (not requiring a cruising phase) and Fig.~\ref{fig:trajsAndProfiles}(d) (requiring a cruising phase) for $L_2$ and  $L_\infty$.

For all these solutions there are several advantages of the $L_2$ solution.  The path requires less time is often a shorter distance.
The path also requires less control changes.  These control changes are shown graphically with $\bullet$ shapes along the $L_2$ path and {\tiny$\blacksquare$} shapes along the $L_\infty$ path.
Additionally, the $L_2$ acceleration and velocity profiles spend more time at their bounds.


\section{Conclusion}
This work found control expressions for position and velocity control in 2D with $L_2$ constraints on acceleration and velocity.
 Future work should extend this to 3D, which is promising for cases with high symmetry such as when the final velocity is zero.
  Extending the work of \cite{carozza2010baserunner} to quickly derive the optimum solution and to incorporate velocity constraints is another exciting direction for future work.
  

\bibliography{biblio}
\bibliographystyle{IEEEtran}

\end{document}


\maketitle

\begin{abstract}
Optimal control solutions are often formulated using $L_\infty$ constraints, even for systems better modeled with $L_2$ constraints.
Given starting and ending positions and velocities, $L_2$ bounds on the acceleration and velocity, and the restriction to no more than two, constant control inputs, this paper provides routines to compute the minimal-time path. Closed form solutions are provided for reaching a position in minimum time with and without a velocity bound.  
Code is provided on GitHub\footnote{ \url{https://github.com/RoboticSwarmControl/2023minTimeL2pathsConstraints/}}.
\end{abstract}

\section{Appendix}

The equations for the two sextic equations are large, and are placed in this appendix to make the paper easier to read.

If the ending configuration is sufficiently far from the initial configuration, the goal is reachable by a two-phase input which consists of a maximum acceleration input in direction $\theta_1$ for $t_1$ seconds, followed by a coasting phase for $t_2$ seconds.

At time $t_1$ the system reaches velocity $v_m$ under a constant acceleration $a_m [\cos(\theta_1 ),\sin(\theta_1 )]^{\top}$:
\begin{align}
\sqrt{(v_{0x}\! +\! a_m \! \cos(\theta_1 )t_1)^2 \! + \! (v_{0y}\! +\! a_m \! \sin(\theta_1 )t_1)^2} = v_m.
\end{align}
This is a quadratic equation with  two solutions for $t_1$, but only the positive value is relevant since we are planning forward in time.
We express $t_1$ as a function of the angle $\theta_1$:
\begin{align}
   t_1(\theta_1) =&\frac{\sqrt{v_m^2 - v_{0x}^2 - v_{0y}^2 +(v_{0x} \cos(\theta_1 )+ v_{0y} \sin (\theta_1 ))^2 }}{a_m} \nonumber \\
   &- \frac{(v_{0x} \cos (\theta_1)+v_{0y} \sin (\theta_1 ))}{a_m}. \label{eq:t1solutioncoast}
\end{align}

The position of the particle at time $t_1$ is
\begin{align}
p_x(t_1) &= p_{0x}+ v_{0x} t_1+\frac{a_m}{2} \cos(\theta_1) t_1^2 \nonumber \\
p_y(t_1) &= p_{0y}+ v_{0y} t_1+\frac{a_m}{2} \sin(\theta_1) t_1^2, \label{eq:positionT1}
\end{align}

\subsection{The system reaches terminal velocity}\label{subsec:reachPointCoasting}

If the ending configuration is sufficiently far from the initial configuration, the goal is reachable by a two-phase input which consists of a maximum acceleration input in direction $\theta_1$ for $t_1$ seconds, followed by a coasting phase for $t_2$ seconds.

At time $t_1$ the system reaches velocity $v_m$ under a constant acceleration $a_m [\cos(\theta_1 ),\sin(\theta_1 )]^{\top}$:
\begin{align}
\sqrt{(v_{0x}\! +\! a_m \! \cos(\theta_1 )t_1)^2 \! + \! (v_{0y}\! +\! a_m \! \sin(\theta_1 )t_1)^2} = v_m.
\end{align}
This is a quadratic equation with  two solutions for $t_1$, but only the positive value is relevant since we are planning forward in time.
We express $t_1$ as a function of the angle $\theta_1$:
\begin{align}
   t_1(\theta_1) =&\frac{\sqrt{v_m^2 - v_{0x}^2 - v_{0y}^2 +( v_{0x} \cos(\theta_1 )+ v_{0y} \sin (\theta_1 ))^2 }}{a_m} \nonumber \\
   &- \frac{(v_{0x} \cos (\theta_1)+v_{0y} \sin (\theta_1 ))}{a_m}. \label{eq:t1solutioncoast}
\end{align}

The position of the particle at time $t_1$ is
\begin{align}
p_x(t_1) &= p_{0x}+ v_{0x} t_1+\frac{a_m}{2} \cos(\theta_1) t_1^2 \nonumber \\
p_y(t_1) &= p_{0y}+ v_{0y} t_1+\frac{a_m}{2} \sin(\theta_1) t_1^2, \label{eq:positionT1}
\end{align}
and the velocity of the particle at time $t_1$ is
\begin{align}
v_x(t_1) &= v_{0x} +a_m \cos(\theta_1) t_1 \nonumber \\
v_y(t_1) &= v_{0y} +a_m \sin(\theta_1) t_1. \qquad \qquad \label{eq:velocityT1}
\end{align}
\begin{figure}[tb]
    \centering
    \begin{overpic}[width=0.24\linewidth]{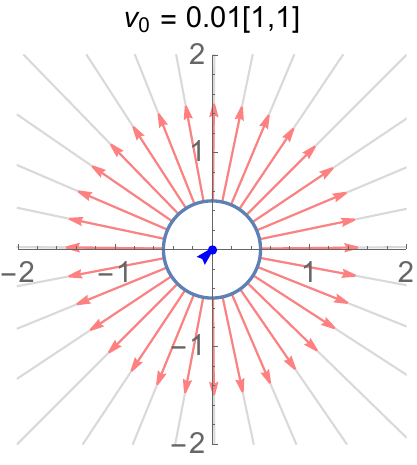}\end{overpic}
    \begin{overpic}[width=0.24\linewidth]{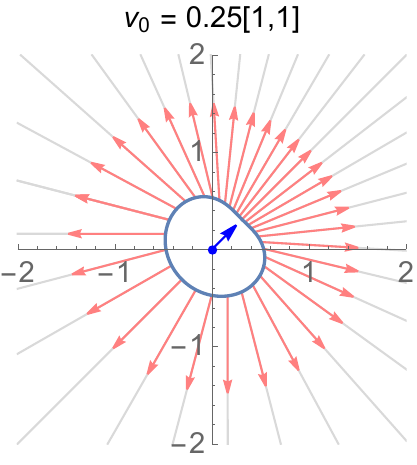}\end{overpic}
    \begin{overpic}[width=0.24\linewidth]{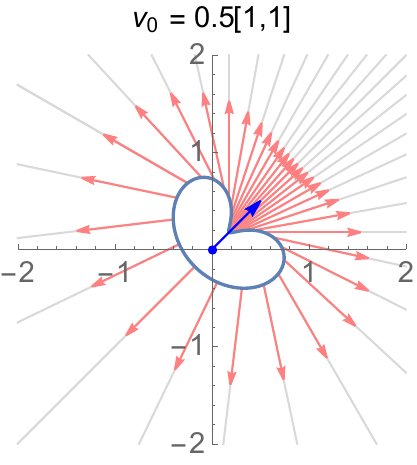}\end{overpic}
    \begin{overpic}[width=0.24\linewidth]{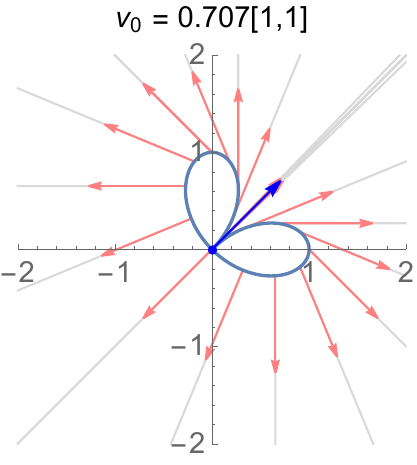}\end{overpic}\\
    \begin{overpic}[width=0.24\linewidth]{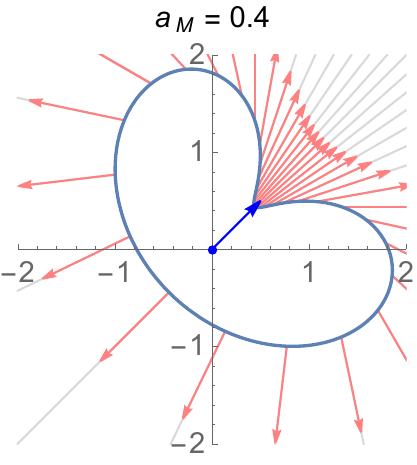}\end{overpic}
    \begin{overpic}[width=0.24\linewidth]{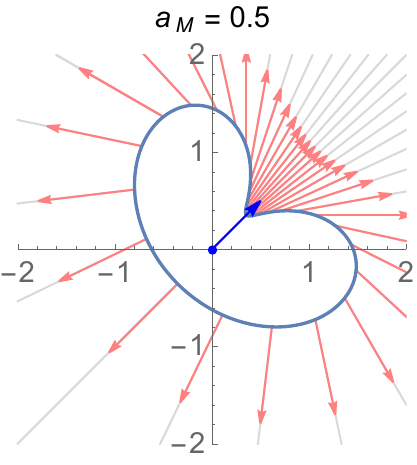}\end{overpic}
    \begin{overpic}[width=0.24\linewidth]{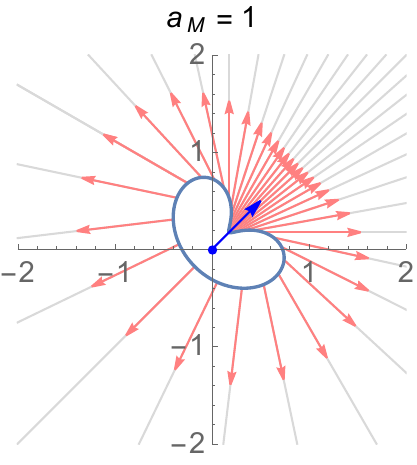}\end{overpic}
    \begin{overpic}[width=0.24\linewidth]{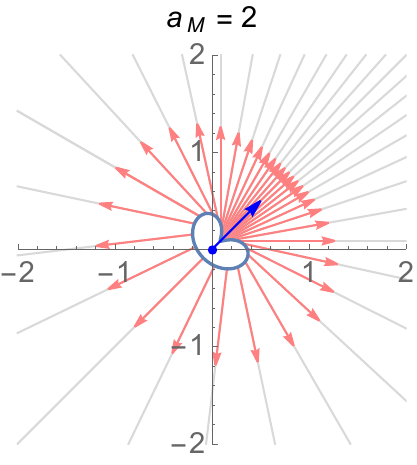}\end{overpic}
    \vspace{-1em}
    \caption{Locus of positions where the particle reaches $v_m = 1$  
    (\adjincludegraphics[width=0.03\columnwidth]{figures/limitLociBlue.pdf}). Top row shows four different starting velocities (blue).  The velocity of the particle due to   thrust $a_m=1$ in directions $\theta \in k \pi /16$, $k\in [0,31]$ is shown with pink arrows, all of length $v_m$.  These arrows point in every direction.
The bottom row shows $\mathbf{v}_0 = [1/2,1/2]$ for four values of $a_m$.  An animation is available at \href{https://youtu.be/2J-p6CDF4FE}{https://youtu.be/2J-p6CDF4FE}.}
    \label{fig:pt1variants}
\end{figure}
Figure~\ref{fig:pt1variants} shows eight variations of the locus of positions at the terminal velocity from \eqref{eq:positionT1} in light blue, along with arrows showing the velocities along this set from \eqref{eq:velocityT1} in pink.
We want solutions for $\theta_1$  that result in the velocity pointing toward to the goal at time $t_1$. 
We could check directly that 
\begin{align}
    \mathrm{arctan}(v_x(t_1),v_y(t_1)) \equiv \mathrm{arctan}(-p_x(t_1),-p_y(t_1)), \label{eq:arctanEquiv}
\end{align}
but this involves solving for inverse trigonometric functions.
Instead, we compare the slope of the velocity to the slope of the position error. 
\begin{align}
    \frac{v_y(t_1)}{v_x(t_1)} \equiv  \frac{-p_y(t_1)}{-p_x(t_1)}  \label{eq:slopeForCoast}
\end{align}
This results in two candidate solutions, but we can check both using \eqref{eq:arctanEquiv} and save the correct solution.
We will also have to check for zeros of the equation in the same way.
The resulting equation is
\begin{align}
    v_x(t_1)p_y(t_1) -v_y(t_1) p_x(t_1)  \equiv 0.  \label{eq:EqualityCoastToPosition}
\end{align}

Solving for $\sin(\theta_1)$ results in a sextic equation. This equation is long, so it is placed in the Appendix~\ref{subsec:AppendixPassGoalTerminalVelocity}. 
We solve for the six roots of \eqref{eq:sCoastingPhaseToReachGoal}, and discard the complex roots.  
Finding the root of a sextic equation can be efficiently computed in many programming languages~\cite{press2007numerical}, see footnotes for examples\footnote{\href{https://numpy.org/doc/stable/reference/generated/numpy.roots.html}{https://numpy.org/doc/stable/reference/generated/numpy.roots.html}}\footnote{\href{https://www.mathworks.com/help/matlab/ref/roots.html}{https://www.mathworks.com/help/matlab/ref/roots.html}}\footnote{\href{https://www.alglib.net/equations/polynomial.php}{https://www.alglib.net/equations/polynomial.php}}.

Each remaining root is a solution for $\sin(\theta_1)$ and provides two possible $\theta_1$ solutions.
We substitute each into \eqref{eq:t1solutioncoast} to get at most 12 candidate $t_1$ solutions.
The smallest, real, non-negative $t_1$  that satisfies
 \eqref{eq:arctanEquiv}  is used.

\subsection{Passing through goal by reaching terminal velocity}\label{subsec:AppendixPassGoalTerminalVelocity}

We can substitute in the solution to $t_1$ from \eqref{eq:t1solutioncoast} into \eqref{eq:EqualityCoastToPosition}, resulting in the inscrutable equality:
\begin{align}
    0 &\scriptscriptstyle 
    \equiv \frac{1}{2 a_m} ( \cos (\theta_1 ) \left(-2 \sin (\theta_1 ) 
    \left(a_m (-p_{0x} v_{0x}+\mathbf{p}_{Gx} v_{0x}+p_{0y} v_{0y}-\mathbf{p}_{Gy} v_{0y}) \right.\right. \nonumber\\
  & \scriptscriptstyle \left.
  + v_{0y}^2 \sqrt{v_m^2-v_{0x}^2+(v_{0x} \cos (\theta_1 )+v_{0y} \sin (\theta_1 ))^2-v_{0y}^2}\right)   \\
&\scriptscriptstyle
+2 a_m (p_{0y}-\mathbf{p}_{Gy}) \sqrt{v_m^2-v_{0x}^2+(v_{0x} \cos (\theta_1 )+v_{0y} \sin (\theta_1 ))^2-v_{0y}^2} \nonumber \\
&\scriptscriptstyle \left.
+v_{0y} \left(v_m^2-2 v_{0x}^2\right)-v_{0y} \cos (2 \theta_1 ) \left(v_{0y}^2-3 v_{0x}^2\right)\right) \nonumber \\
&\scriptscriptstyle 
-\sin (\theta_1 ) \left(2 a_m (p_{0x}-\mathbf{p}_{Gx}) \sqrt{v_m^2-v_{0x}^2+(v_{0x} \cos (\theta_1 )+v_{0y} \sin (\theta_1 ))^2-v_{0y}^2} \right. \nonumber \\
&\scriptscriptstyle \left.
+v_m^2 v_{0x} +v_{0x} \cos (2 \theta_1 ) (v_{0x}-v_{0y}) (v_{0x}+v_{0y}) \right) \nonumber \\
&\scriptscriptstyle
+2 v_{0y} \cos ^2(\theta_1 ) (a_m (\mathbf{p}_{Gx}-p_{0x})+2 v_{0x} v_{0y} \sin (\theta_1 )) + 2 a_m v_{0x} \sin ^2(\theta_1 ) (p_{0y}-\mathbf{p}_{Gy})
\nonumber \\
&\scriptscriptstyle
+v_{0x} (v_{0x} \sin (2 \theta_1 )-2 v_{0y} \cos (2 \theta_1 )) \sqrt{v_m^2-v_{0x}^2+(v_{0x} \cos (\theta_1 )+v_{0y} \sin (\theta ))^2-v_{0y}^2}). \nonumber
\end{align}

This function is still to hard to solve, so we need to make some substitutions.
First, we rotate our coordinate frame so that $v_{0y} = 0$.
Next, we apply a change of variables to eliminate the two trigonometric functions: $\sin (\theta_1 )= s$, and $\cos (\theta_1)=\pm \sqrt{1-s^2}$.
This results in the more tractable equation
\begin{align}
 &\scriptscriptstyle 
    s v_{0x} \left(v_m^2+v_{0x} \left(v_{0x} -2 \sqrt{1-s^2} \sqrt{(v_m-s v_{0x}) (v_m+s v_{0x})}-2 s^2 v_{0x}\right)\right) \nonumber \\
&\scriptscriptstyle
    \equiv \quad 2 a_m \left(p_{0x} \sqrt{1-s^2} s v_{0x} -p_{0x} s \sqrt{(v_m-s v_{0x}) (v_m+s v_{0x})} \right. \nonumber \\
&\scriptscriptstyle \left.    
   \quad +p_{0y} \sqrt{1-s^2} \sqrt{(v_m-s v_{0x}) (v_m+s v_{0x})} +p_{0y} s^2 v_{0x}\right) \label{eq:solveForsCoast}.
\end{align}
 Rewriting \eqref{eq:solveForsCoast} gives \eqref{eq:sCoastingPhaseToReachGoal},  a sextic equation in $s$. 
\begin{align}
0 =  16 a_m^4 v_m^4 p_{0y}^4 \nonumber\\
+ 64 a_m^3 v_m^4 p_{0y}^3  v_{0x}^2  \textcolor{red}{s} \nonumber\\
 \nonumber\\
 - 8 a_m^2 v_m^2 p_{0y}^2  (4 a_m^2 v_m^2 p_{0x}^2 + 4 a_m^2 v_m^2 p_{0y}^2 +  \nonumber\\
    v_m^4 v_{0x}^2 + 4 a_m^2 p_{0x}^2 v_{0x}^2 + 4 a_m^2 p_{0y}^2 v_{0x}^2 -  \nonumber\\
    10 v_m^2 v_{0x}^4 + v_{0x}^6) \textcolor{red}{s^2} \nonumber\\
    \nonumber\\
  - 16 a_m v_m^2 p_{0y} v_{0x}^2 (6 a_m^2 v_m^2 p_{0y}^2 + v_m^4 v_{0x}^2 + \nonumber\\
    6 a_m^2 p_{0y}^2 v_{0x}^2 - 2 v_m^2 v_{0x}^4 + v_{0x}^6) \textcolor{red}{s^3} \nonumber\\
    \nonumber\\
 + (16 a_m^4 v_m^4 p_{0x}^4 + 32 a_m^4 v_m^4 p_{0x}^2 p_{0y}^2 + \nonumber\\
    16 a_m^4 v_m^4 p_{0y}^4 - 8 a_m^2 v_m^6 p_{0x}^2 v_{0x}^2 - \nonumber\\
    32 a_m^4 v_m^2 p_{0x}^4 v_{0x}^2 + 8 a_m^2 v_m^6 p_{0y}^2 v_{0x}^2 + \nonumber\\
    32 a_m^4 v_m^2 p_{0y}^4 v_{0x}^2 + v_m^8 v_{0x}^4 + 8 a_m^2 v_m^4 p_{0x}^2 v_{0x}^4 + \nonumber\\
    16 a_m^4 p_{0x}^4 v_{0x}^4 - 72 a_m^2 v_m^4 p_{0y}^2 v_{0x}^4 + \nonumber\\
    32 a_m^4 p_{0x}^2 p_{0y}^2 v_{0x}^4 + 16 a_m^4 p_{0y}^4 v_{0x}^4 - 4 v_m^6 v_{0x}^6 + \nonumber\\
    8 a_m^2 v_m^2 p_{0x}^2 v_{0x}^6 - 72 a_m^2 v_m^2 p_{0y}^2 v_{0x}^6 + \nonumber\\
    6 v_m^4 v_{0x}^8 - 8 a_m^2 p_{0x}^2 v_{0x}^8 + 8 a_m^2 p_{0y}^2 v_{0x}^8 - \nonumber\\
    4 v_m^2 v_{0x}^{10} + v_{0x}^{12}) \textcolor{red}{s^4} \nonumber\\
    \nonumber\\
+  8 a_m p_{0y} v_{0x}^2 (4 a_m^2 v_m^4 p_{0x}^2 + 4 a_m^2 v_m^4 p_{0y}^2 + \nonumber\\
    v_m^6 v_{0x}^2 - 8 a_m^2 v_m^2 p_{0x}^2 v_{0x}^2 + 8 a_m^2 v_m^2 p_{0y}^2 v_{0x}^2 - \nonumber\\
    v_m^4 v_{0x}^4 + 4 a_m^2 p_{0x}^2 v_{0x}^4 + 4 a_m^2 p_{0y}^2 v_{0x}^4 - \nonumber\\
    v_m^2 v_{0x}^6 + v_{0x}^8)  \textcolor{red}{s^5} \nonumber\\
        \nonumber\\
+  16 a_m^2  v_{0x}^4 (v_m^4 p_{0x}^2 + v_m^4 p_{0y}^2 - 2 v_m^2 p_{0x}^2 v_{0x}^2 + \nonumber\\
    2 v_m^2 p_{0y}^2 v_{0x}^2 + p_{0x}^2 v_{0x}^4 + p_{0y}^2 v_{0x}^4) \textcolor{red}{s^6}  \label{eq:sCoastingPhaseToReachGoal}
\end{align}

\subsection{Stopping at goal no terminal velocity}\label{subsec:StopGoalNoTerminalVelocity}

If the starting and ending position are sufficiently close such that the velocity never exceeds $v_m$, then the goal is reachable in minimum time by a two-phase input which consists of a maximum acceleration input in direction $\theta_1$ for $t_1$ seconds, followed by a maximum acceleration input opposing the current velocity to bring the system to rest in $t_3 = \mathbf{v}(t_1)/a_m$ seconds ($t_2 = 0$).

 After applying the constant input $a_m[\cos(\theta_1), \sin(\theta_1)]^{\top}$ for $t_1$ seconds, the position and velocity are
 \begin{align}
     \mathbf{p}(t_1) &= \begin{bmatrix}
   & p_{0x}+ v_{0x} t_1            + &\frac{a_m}{2} \cos (\theta_1 ) t_1^2\\
   & p_{0y}\phantom{+t_1 v_{0y}} + &\frac{a_m}{2} \sin (\theta_1 ) t_1^2
     \end{bmatrix} \nonumber \\
     \mathbf{v}(t_1) &= \begin{bmatrix}
   & v_{0x}+ & a_m \cos (\theta_1 ) t_1\\
   &         & a_m \sin (\theta_1 ) t_1
     \end{bmatrix}
     \label{eq:PosAndVelTheta1Time1}
 \end{align}
 The deceleration command is in the opposite direction of $\mathbf{v}(t_1)$ so that $\theta_3 = \arctan(-\mathbf{v}_x(t_1),-\mathbf{v}_y(t_1))$, and lasts for $t_3 = \norm{\mathbf{v}(t_1)}/a_m $ seconds.
 At time $t_1+t_3$, we want the $x$ and $y$ positions to be zero and the final velocity to be zero.
 The final position is entirely controlled by the initial conditions and the selected $\theta_1$ and $\theta_3$:
 \begin{align}
   t_3 &=\frac{\norm{\mathbf{v}(t_1)}}{a_m} = \sqrt{\left( \frac{v_{0x}}{a_m} + \cos(\theta_1 ) t_1 \right)^2 + (\sin (\theta_1 ) t_1 )^2}\nonumber\\
   0 &= 
  p_{x}(t_1)  + \mathbf{v}_{x}(t_1) \frac{t_3}{2}  \nonumber \\
    0 &=
  p_{y}(t_1)  + \mathbf{v}_{y}(t_1)  \frac{t_3}{2}. \label{eq:posForZeroFinalVelocity}
 \end{align}
We then scale the starting position and velocity by dividing each by $a_m$ and remove the term $a_m$ from the calculation:  $\tilde{\mathbf{p}}_{0} = \mathbf{p}_{0}/a_m$,
$\tilde{\mathbf{v}}_{0} = \mathbf{v}_{0}/a_m$.
 We apply a change of variables to eliminate the two trigonometric functions: $\cos (\theta_1 )= c$, and $\sin (\theta_1)=\pm \sqrt{1-c^2}$.
 The resulting position constraints simplify to:
 \begin{align}
    0 &= 2 \tilde{p_{0x}} + 2 \tilde{v_{0x}} t_1 +c t_1^2 + (\tilde{v_{0x}}\! +\! c t_1) \sqrt{\tilde{v_{0x}}^2 +\! 2 c  \tilde{v_{0x}} t_1 +\!t_1^2} \nonumber\\
    0 &= 2 \tilde{p_{0y}} + \sqrt{1-c^2} t_1 \left(\sqrt{\tilde{v_{0x}}^2 + 2 c  \tilde{v_{0x}} t_1 + t_1^2}+t_1\!\right).\! \label{eq:positionUsingThrustToStopAtGoalNoTermalVelocity}
 \end{align}
 This set of equations can be solved for $c$ as a function of $t_1$.
The calculations are long, but they are included here for completeness.

The set of equations \eqref{eq:positionUsingThrustToStopAtGoalNoTermalVelocity} can be solved for $c$ as a function of $t_1$.
Like \eqref{eq:solveForsCoast}, the resulting equation is sextic, but this time in 
 $\textcolor{red}{t_1}$:
\begin{align}\label{eq:bbtequation}
  0 =&\scriptscriptstyle   -4 (16 \tilde{p_{0x}}^6 + 8 \tilde{p_{0x}}^4 (6 \tilde{p_{0y}}^2 - \tilde{v_{0x}}^4) + ( \tilde{p_{0y}} \tilde{v_{0x}}^4 -4 \tilde{p_{0y}}^3)^2 + 
   \tilde{p_{0x}}^2 (48 \tilde{p_{0y}}^4 + 48 \tilde{p_{0y}}^2 \tilde{v_{0x}}^4 + \tilde{v_{0x}}^8))  \nonumber \\
 &\scriptscriptstyle + \textcolor{red}{t_1} (-4 \tilde{p_{0x}} \tilde{v_{0x}} (80 \tilde{p_{0x}}^4 + 80 \tilde{p_{0y}}^4 + 72 \tilde{p_{0y}}^2 \tilde{v_{0x}}^4 + \tilde{v_{0x}}^8  
   8 \tilde{p_{0x}}^2 (20 \tilde{p_{0y}}^2 - 3 \tilde{v_{0x}}^4)))  \nonumber \\
 &\scriptscriptstyle + \textcolor{red}{t_1^2} (   -\tilde{v_{0x}}^2 (464 \tilde{p_{0x}}^4 + 208 \tilde{p_{0y}}^4 + 88 \tilde{p_{0y}}^2 \tilde{v_{0x}}^4 + \tilde{v_{0x}}^8 + 
   24 \tilde{p_{0x}}^2 (28 \tilde{p_{0y}}^2 - 5 \tilde{v_{0x}}^4))) \nonumber \\
 &\scriptscriptstyle + \textcolor{red}{t_1^3} (-256 \tilde{p_{0x}}^3 \tilde{v_{0x}}^3 - 128 \tilde{p_{0x}} \tilde{p_{0y}}^2 \tilde{v_{0x}}^3 + 64 \tilde{p_{0x}} \tilde{v_{0x}}^7)  \nonumber \\
 &\scriptscriptstyle + \textcolor{red}{t_1^4} (64 \tilde{p_{0x}}^4 + 128 \tilde{p_{0x}}^2 \tilde{p_{0y}}^2 + 64 \tilde{p_{0y}}^4 - 64 \tilde{p_{0x}}^2 \tilde{v_{0x}}^4 + 
    32 \tilde{p_{0y}}^2 \tilde{v_{0x}}^4 + 12 \tilde{v_{0x}}^8)  \nonumber \\
 &\scriptscriptstyle + \textcolor{red}{t_1^5} (64 \tilde{p_{0x}}^3 \tilde{v_{0x}} + 64 \tilde{p_{0x}} \tilde{p_{0y}}^2 \tilde{v_{0x}} - 16 \tilde{p_{0x}} \tilde{v_{0x}}^5)  \nonumber \\
 & \scriptscriptstyle + \textcolor{red}{t_1^6} (16 \tilde{p_{0x}}^2 \tilde{v_{0x}}^2 - 4 \tilde{v_{0x}}^6). 
 \end{align}
 The equation for $c(t_1)$ is inscrutable, but is composed of constants. The variable $\textcolor{red}{t_1}$ appears five times:
\begin{equation}
\begin{aligned}[b]
    c(\textcolor{red}{t_1}) = \hspace{33 em} \\
    (4 \tilde{p_{0x}} \tilde{v_{0x}}^2 (-4096 (\tilde{p_{0x}}^2 + \tilde{p_{0y}}^2)^5 (15 \tilde{p_{0x}}^4 - 49 \tilde{p_{0x}}^2 \tilde{p_{0y}}^2 + 40 \tilde{p_{0y}}^4) + \\
       2048 (\tilde{p_{0x}}^2 + \tilde{p_{0y}}^2)^2 (49 \tilde{p_{0x}}^8 - 41 \tilde{p_{0x}}^6 \tilde{p_{0y}}^2 -  303 \tilde{p_{0x}}^4 \tilde{p_{0y}}^4 + 529 \tilde{p_{0x}}^2 \tilde{p_{0y}}^6 - 186 \tilde{p_{0y}}^8) \tilde{v_{0x}}^4 - \\
       256 (273 \tilde{p_{0x}}^{10} - 28 \tilde{p_{0x}}^8 \tilde{p_{0y}}^2 - 1114 \tilde{p_{0x}}^6 \tilde{p_{0y}}^4 +   804 \tilde{p_{0x}}^4 \tilde{p_{0y}}^6 \\
       - 1263 \tilde{p_{0x}}^2 \tilde{p_{0y}}^8 + 1216 \tilde{p_{0y}}^{10}) \tilde{v_{0x}}^8 + \\
       256 (105 \tilde{p_{0x}}^8 - 183 \tilde{p_{0x}}^6 \tilde{p_{0y}}^2 + 163 \tilde{p_{0x}}^4 \tilde{p_{0y}}^4 -   73 \tilde{p_{0x}}^2 \tilde{p_{0y}}^6 - 220 \tilde{p_{0y}}^8) \tilde{v_{0x}}^{12} + \\
       16 (-385 \tilde{p_{0x}}^6 + 1278 \tilde{p_{0x}}^4 \tilde{p_{0y}}^2 - 2233 \tilde{p_{0x}}^2 \tilde{p_{0y}}^4 +    1192 \tilde{p_{0y}}^6) \tilde{v_{0x}}^{16} + \\
       8 (105 \tilde{p_{0x}}^4 - 443 \tilde{p_{0x}}^2 \tilde{p_{0y}}^2 + 542 \tilde{p_{0y}}^4) \tilde{v_{0x}}^{20} +    7 (-9 \tilde{p_{0x}}^2 + 32 \tilde{p_{0y}}^2) \tilde{v_{0x}}^{24} + 2 \tilde{v_{0x}}^{28}) + \\
    \textcolor{red}{t_1} (-4096 (\tilde{p_{0x}}^2 + \tilde{p_{0y}}^2)^4 (159 \tilde{p_{0x}}^6 - 344 \tilde{p_{0x}}^4 \tilde{p_{0y}}^2 +      51 \tilde{p_{0x}}^2 \tilde{p_{0y}}^4 + 74 \tilde{p_{0y}}^6) \tilde{v_{0x}}^3 + \\
       2048 (\tilde{p_{0x}}^2 + \tilde{p_{0y}}^2) (497 \tilde{p_{0x}}^{10} - 79 \tilde{p_{0x}}^8 \tilde{p_{0y}}^2 -   1890 \tilde{p_{0x}}^6 \tilde{p_{0y}}^4 + 1510 \tilde{p_{0x}}^4 \tilde{p_{0y}}^6 + 673 \tilde{p_{0x}}^2 \tilde{p_{0y}}^8 - \\
          359 \tilde{p_{0y}}^{10}) \tilde{v_{0x}}^7 - \\
       256 (2593 \tilde{p_{0x}}^{10} - 360 \tilde{p_{0x}}^8 \tilde{p_{0y}}^2 - 7914 \tilde{p_{0x}}^6 \tilde{p_{0y}}^4 +  6700 \tilde{p_{0x}}^4 \tilde{p_{0y}}^6 \\
       - 2223 \tilde{p_{0x}}^2 \tilde{p_{0y}}^8 + 1988 \tilde{p_{0y}}^{10}) \tilde{v_{0x}}^{11} + \\
       256 (905 \tilde{p_{0x}}^8 - 862 \tilde{p_{0x}}^6 \tilde{p_{0y}}^2 - 1020 \tilde{p_{0x}}^4 \tilde{p_{0y}}^4 +     1106 \tilde{p_{0x}}^2 \tilde{p_{0y}}^6 - 201 \tilde{p_{0y}}^8) \tilde{v_{0x}}^{15} + \\
       16 (-2865 \tilde{p_{0x}}^6 + 4204 \tilde{p_{0x}}^4 \tilde{p_{0y}}^2 - 1685 \tilde{p_{0x}}^2 \tilde{p_{0y}}^4 +      1886 \tilde{p_{0y}}^6) \tilde{v_{0x}}^{19} + \\
       8 (617 \tilde{p_{0x}}^4 - 966 \tilde{p_{0x}}^2 \tilde{p_{0y}}^2 +    709 \tilde{p_{0y}}^4) \tilde{v_{0x}}^{23} + (-239 \tilde{p_{0x}}^2 + 272 \tilde{p_{0y}}^2) \tilde{v_{0x}}^{27} +  2 \tilde{v_{0x}}^{31} + \\
       \textcolor{red}{t_1} (8 \tilde{p_{0x}} (-4096 (\tilde{p_{0x}}^2 - 2 \tilde{p_{0y}}^2) (\tilde{p_{0x}}^2 + \tilde{p_{0y}}^2)^6 -    1024 (\tilde{p_{0x}}^2 + \tilde{p_{0y}}^2)^3 (44 \tilde{p_{0x}}^6 - 81 \tilde{p_{0x}}^4 \tilde{p_{0y}}^2 - \\
                10 \tilde{p_{0x}}^2 \tilde{p_{0y}}^4 + 19 \tilde{p_{0y}}^6) \tilde{v_{0x}}^4 +   256 (299 \tilde{p_{0x}}^{10} + 45 \tilde{p_{0x}}^8 \tilde{p_{0y}}^2 - 1018 \tilde{p_{0x}}^6 \tilde{p_{0y}}^4 + 434 \tilde{p_{0x}}^4 \tilde{p_{0y}}^6\\
                 + 559 \tilde{p_{0x}}^2 \tilde{p_{0y}}^8 - 127 \tilde{p_{0y}}^{10}) \tilde{v_{0x}}^8 -  128 (400 \tilde{p_{0x}}^8 - 317 \tilde{p_{0x}}^6 \tilde{p_{0y}}^2 - 715 \tilde{p_{0x}}^4 \tilde{p_{0y}}^4 + \\
                753 \tilde{p_{0x}}^2 \tilde{p_{0y}}^6 - 17 \tilde{p_{0y}}^8) \tilde{v_{0x}}^{12} +       16 (1125 \tilde{p_{0x}}^6 - 1602 \tilde{p_{0x}}^4 \tilde{p_{0y}}^2 - 275 \tilde{p_{0x}}^2 \tilde{p_{0y}}^4 + 
                852 \tilde{p_{0y}}^6) \tilde{v_{0x}}^{16}
                \\
                 -       4 (884 \tilde{p_{0x}}^4 - 1401 \tilde{p_{0x}}^2 \tilde{p_{0y}}^2 + 255 \tilde{p_{0y}}^4) \tilde{v_{0x}}^{20} +  9 (41 \tilde{p_{0x}}^2 - 47 \tilde{p_{0y}}^2) \tilde{v_{0x}}^{24} - 16 \tilde{v_{0x}}^{28}) + \\
          4 \textcolor{red}{t_1} \tilde{v_{0x}} (4096 (\tilde{p_{0x}}^2 + \tilde{p_{0y}}^2)^5 (4 \tilde{p_{0x}}^4 - 11 \tilde{p_{0x}}^2 \tilde{p_{0y}}^2 + 6 \tilde{p_{0y}}^4) - \\
             1024 (\tilde{p_{0x}}^2 + \tilde{p_{0y}}^2)^2 (45 \tilde{p_{0x}}^8 - 31 \tilde{p_{0x}}^6 \tilde{p_{0y}}^2 -       167 \tilde{p_{0x}}^4 \tilde{p_{0y}}^4 + 219 \tilde{p_{0x}}^2 \tilde{p_{0y}}^6 - 58 \tilde{p_{0y}}^8) \tilde{v_{0x}}^4 + \\
             256 (189 \tilde{p_{0x}}^{10} + 6 \tilde{p_{0x}}^8 \tilde{p_{0y}}^2 - 702 \tilde{p_{0x}}^6 \tilde{p_{0y}}^4 +    732 \tilde{p_{0x}}^4 \tilde{p_{0y}}^6 - 359 \tilde{p_{0x}}^2 \tilde{p_{0y}}^8 + \\
                182 \tilde{p_{0y}}^{10}) \tilde{v_{0x}}^8 -  128 (205 \tilde{p_{0x}}^8 - 217 \tilde{p_{0x}}^6 \tilde{p_{0y}}^2 - 290 \tilde{p_{0x}}^4 \tilde{p_{0y}}^4 +  447 \tilde{p_{0x}}^2 \tilde{p_{0y}}^6 \\
                - 109 \tilde{p_{0y}}^8) \tilde{v_{0x}}^{12} +   16 (510 \tilde{p_{0x}}^6 - 883 \tilde{p_{0x}}^4 \tilde{p_{0y}}^2 + 255 \tilde{p_{0x}}^2 \tilde{p_{0y}}^4 + 56 \tilde{p_{0y}}^6) \tilde{v_{0x}}^{16}\\
                 -    4 (369 \tilde{p_{0x}}^4 - 681 \tilde{p_{0x}}^2 \tilde{p_{0y}}^2 +    308 \tilde{p_{0y}}^4) \tilde{v_{0x}}^{20} + (145 \tilde{p_{0x}}^2 - 188 \tilde{p_{0y}}^2) \tilde{v_{0x}}^{24} -   6 \tilde{v_{0x}}^{28} + \\
           4 \textcolor{red}{t_1} \tilde{p_{0x}} \tilde{v_{0x}} (4 (\tilde{p_{0x}}^2 + \tilde{p_{0y}}^2) -    \tilde{v_{0x}}^4) (512 (\tilde{p_{0x}}^2 + \tilde{p_{0y}}^2)^3 (3 \tilde{p_{0x}}^4 -\\ 
                   7 \tilde{p_{0x}}^2 \tilde{p_{0y}}^2 + 3 \tilde{p_{0y}}^4) -       64 (32 \tilde{p_{0x}}^8 - 13 \tilde{p_{0x}}^6 \tilde{p_{0y}}^2 - 119 \tilde{p_{0x}}^4 \tilde{p_{0y}}^4 + \\
                   169 \tilde{p_{0x}}^2 \tilde{p_{0y}}^6 - 45 \tilde{p_{0y}}^8) \tilde{v_{0x}}^4 +    16 (68 \tilde{p_{0x}}^6 - 95 \tilde{p_{0x}}^4 \tilde{p_{0y}}^2 - 40 \tilde{p_{0x}}^2 \tilde{p_{0y}}^4 + \\
                   75 \tilde{p_{0y}}^6) \tilde{v_{0x}}^8 -      4 (72 \tilde{p_{0x}}^4 - 135 \tilde{p_{0x}}^2 \tilde{p_{0y}}^2 + 61 \tilde{p_{0y}}^4) \tilde{v_{0x}}^{12} + \\
                19 (2 \tilde{p_{0x}}^2 - 3 \tilde{p_{0y}}^2) \tilde{v_{0x}}^{16} - 2 \tilde{v_{0x}}^{20}) +    \textcolor{red}{t_1^2} \tilde{v_{0x}}^2 (-2 \tilde{p_{0x}} + \tilde{v_{0x}}^2) (2 \tilde{p_{0x}} + \\
                \tilde{v_{0x}}^2) (-256 (\tilde{p_{0x}}^2 + \tilde{p_{0y}}^2)^3 (7 \tilde{p_{0x}}^4 -          15 \tilde{p_{0x}}^2 \tilde{p_{0y}}^2 + 6 \tilde{p_{0y}}^4) + \\
                128 (18 \tilde{p_{0x}}^8 - 11 \tilde{p_{0x}}^6 \tilde{p_{0y}}^2 - 51 \tilde{p_{0x}}^4 \tilde{p_{0y}}^4 +          83 \tilde{p_{0x}}^2 \tilde{p_{0y}}^6 - 23 \tilde{p_{0y}}^8) \tilde{v_{0x}}^4 - \\
                32 (37 \tilde{p_{0x}}^6 - 59 \tilde{p_{0x}}^4 \tilde{p_{0y}}^2 - 11 \tilde{p_{0x}}^2 \tilde{p_{0y}}^4 +          37 \tilde{p_{0y}}^6) \tilde{v_{0x}}^8 + \\
                8 (38 \tilde{p_{0x}}^4 - 73 \tilde{p_{0x}}^2 \tilde{p_{0y}}^2 +           31 \tilde{p_{0y}}^4) \tilde{v_{0x}}^{12} + (-39 \tilde{p_{0x}}^2 + 56 \tilde{p_{0y}}^2) \tilde{v_{0x}}^{16} +        2 \tilde{v_{0x}}^{20} \\
                )))))/(8 (4 (\tilde{p_{0x}}^2 + \tilde{p_{0y}}^2) -  \tilde{v_{0x}}^4) (16 (\tilde{p_{0x}}^2 + \tilde{p_{0y}}^2)^3 - \\
      8 (\tilde{p_{0x}}^4 - 6 \tilde{p_{0x}}^2 \tilde{p_{0y}}^2 + \tilde{p_{0y}}^4) \tilde{v_{0x}}^4 + (\tilde{p_{0x}}^2 +      \tilde{p_{0y}}^2) \tilde{v_{0x}}^8) (64 \tilde{p_{0x}}^8 + 16 \tilde{p_{0x}}^6 (4 \tilde{p_{0y}}^2 + 5 \tilde{v_{0x}}^4) - \\
      2 (8 \tilde{p_{0y}}^4 + 10 \tilde{p_{0y}}^2 \tilde{v_{0x}}^4 + \tilde{v_{0x}}^8)^2 -   12 \tilde{p_{0x}}^4 (16 \tilde{p_{0y}}^4 + 56 \tilde{p_{0y}}^2 \tilde{v_{0x}}^4 + 7 \tilde{v_{0x}}^8) + \\
      \tilde{p_{0x}}^2 (-320 \tilde{p_{0y}}^6 + 976 \tilde{p_{0y}}^4 \tilde{v_{0x}}^4 + 324 \tilde{p_{0y}}^2 \tilde{v_{0x}}^8 +     23 \tilde{v_{0x}}^{12}))) 
              \end{aligned}
\label{eq:bbtequationFort}
\end{equation}
\bibliography{biblio}
\bibliographystyle{IEEEtran}